\documentclass{article}

\PassOptionsToPackage{numbers,sort}{natbib}

\usepackage[preprint]{styles/neurips_2024}

\usepackage[utf8]{inputenc} 
\usepackage[T1]{fontenc}    
\usepackage[colorlinks,citecolor=linkColor]{hyperref}       
\usepackage{url}            
\usepackage{booktabs}       
\usepackage{amsfonts}       
\usepackage{nicefrac}       
\usepackage{microtype}      
\usepackage{xcolor}         
\usepackage{colortbl}
\usepackage{graphicx} 
\usepackage{wrapfig}
\usepackage{multirow}
\usepackage{xspace}
\usepackage{pifont}
\usepackage{enumerate}
\usepackage{amsmath}
\usepackage{subcaption}
\UseRawInputEncoding
\def\ie{\emph{i.e.}}
\def\eg{\emph{e.g.}}

\def\miniinternvl{Mini-InternVL\xspace} 
\def\internvlad{\miniinternvl-DA\xspace} 
\def\internvlrs{\miniinternvl-DA\xspace} 
\def\internvlmi{\miniinternvl-DA\xspace}
\def\internvlbdd{\miniinternvl-DA\xspace} 
\newcommand{\yes}{\ding{51}}
\newcommand{\no}{\ding{55}}

\newcommand{\figref}[1]{Figure~\ref{#1}}
\newcommand{\tabref}[1]{Table~\ref{#1}}

\newcommand{\secref}[1]{Section~\ref{#1}}

\definecolor{linkColor}{rgb}{0.18,0.39,0.62}

\title{Mini-InternVL: A Flexible-Transfer Pocket Multimodal Model with 5\% Parameters and 90\% Performance}

\author{%
\textbf{
Zhangwei Gao$^{1,7*}$ \quad
Zhe Chen$^{1,3*}$ \quad
Erfei Cui$^{1,7*}$ \quad
Yiming Ren$^{1,2*}$ \quad
Weiyun Wang$^{1,4}$\thanks{Equal contributions}}
\\
\textbf{
Jinguo Zhu$^{1}$ \quad
Hao Tian$^{6}$ \quad
Shenglong Ye$^{1}$ \quad
Junjun He$^{1}$ \quad
Xizhou Zhu$^{2,1}$ \quad
Lewei Lu$^{6}$  \quad }
\\
\textbf{
Tong Lu$^{3}$ \quad
Yu Qiao$^{1}$ \quad
Jifeng Dai$^{2,1}$ \quad
Wenhai Wang$^{5,1}$\thanks{Corresponding Author}}
\\
\\
$^{1}$Shanghai AI Laboratory, $^{2}$Tsinghua University, $^{3}$Nanjing University,\\
$^{4}$Fudan University, $^{5}$The Chinese University of Hong Kong, \\
$^{6}$SenseTime Research, $^{7}$Shanghai Jiao Tong University
\\
\\
\url{https://github.com/OpenGVLab/InternVL}
}

\begin{document}
\maketitle

\begin{abstract}
Multimodal large language models (MLLMs) have demonstrated impressive performance in vision-language tasks across a broad spectrum of domains. However, the large model scale and associated high computational costs pose significant challenges for training and deploying MLLMs on consumer-grade GPUs or edge devices, thereby hindering their widespread application. In this work, we introduce Mini-InternVL, a series of MLLMs with parameters ranging from 1B to 4B, which achieves 90\% of the performance with only 5\% of the parameters. This significant improvement in efficiency and effectiveness makes our models more accessible and applicable in various real-world scenarios. To further promote the adoption of our models, we develop a unified adaptation framework for \miniinternvl, which enables our models to transfer and outperform specialized models in downstream tasks, including autonomous driving, medical images, and remote sensing. We believe that our study can provide valuable insights and resources to advance the development of efficient and effective MLLMs.
\end{abstract}

\section{Introduction}
\label{sec:intr}

In recent years, there have been significant advancements in multimodal large language models (MLLMs) \cite{chen2023internvl,chen2024far,gpt4v,openai2024gpt4o,li2024llava,liu2024llavanext, wang2024emu3,lin2024vila,ye2024mplug,ge2024convllava,shi2024eagle}, which leverages the powerful capabilities of pre-trained large language models (LLMs) \cite{llama3modelcard,touvron2023llama2,cai2024internlm2,qwen,qwen2,jiang2023mistral} alongside vision foundation models (VFMs)~\cite{chen2023internvl,radford2021learning,zhai2023sigmoid}. These models undergo multi-stage training on extensive image-text data, which effectively aligns visual representations from VFMs with the latent space of LLMs, leading to promising performance in general vision-language understanding, reasoning, and interaction tasks. However, \emph{the large computational burden and the poor performance on long-tail domain-specific tasks hinder the widespread application of MLLMs in practical scenarios.}

The emergence of lightweight MLLMs~\cite{yao2024minicpm,li2024mgm,chu2023mobilevlm,chu2024mobilevlm,beyer2024paligemma,he2024bunny} has provided a good balance between parameter size and performance, alleviating the reliance on expensive computing devices and fostering the development of various downstream applications. However, there are still several challenges: (1) Most existing MLLMs use vision encoders like CLIP~\cite{radford2021learning}, which are trained on Internet-domain image-text data and are aligned with BERT~\cite{devlin2018bert, jia2021scaling}. \emph{As a result, these vision encoders are not capable of covering the extensive range of visual domains and are misaligned with LLMs' representations.
} 
(2) To adapt MLLMs to specialized domains, existing methods mainly focus on modifying the model architectures, gathering extensive related training data, or customizing the training process for the target domain. 
\emph{There is still no consensus framework for LLMs' downstream adaptation. Different domains have different model designs, data formats, and training schedules.} 

To address these issues, there is a need for a strong vision encoder with comprehensive visual knowledge as well as a general transfer learning paradigm that allows for efficient application across downstream tasks in various domains at a low marginal cost.

In this work, we introduce \miniinternvl, a series of powerful pocket-sized MLLMs that can be easily transferred to various specialized domains. To this end, we first enhance the representational capabilities of a lightweight vision encoder.
We initialize a 300M vision encoder using the weights from CLIP and apply knowledge distillation using InternViT-6B~\cite{chen2023internvl} as the teacher model.
Subsequently, we develop Mini-InternVL series with 1 billion, 2 billion, and 4 billion parameters, by integrating the vision encoder with the pre-trained LLMs such as Qwen2-0.5B~\cite{qwen2}, InternLM2-1.8B~\cite{cai2024internlm2}, and Phi-3-Mini~\cite{abdin2024phi}, respectively. 
Benefiting from the robust vision encoder, Mini-InternVL exhibits excellent multimodal performance on general multimodal benchmarks like MMBench~\cite{liu2023mmbench}, ChartQA~\cite{masry2022chartqa}, and MathVista~\cite{lu2023mathvista}. Remarkably, compared with InternVL2-76B, the proposed Mini-InternVL-4B achieves 90\% of the performance of larger counterparts while using only 5\% of the parameters, significantly reducing computational overhead.

To further adapt our models to specific-domain downstream tasks, we introduce a straightforward yet effective transfer learning paradigm. Within this paradigm, we develop a unified transfer approach applicable to various downstream tasks, including autonomous driving, medical images, and remote sensing. This approach standardizes the model architecture, data format, and training schedule. The results demonstrate the effectiveness of this method in enhancing the models' visual understanding and reasoning capabilities in domain-specific scenarios, enabling them to match the performance of proprietary commercial models within the target domains.

In summary, our contribution has three folds:

(1) We propose Mini-InternVL, a powerful pocket multimodal model, that not only achieves robust multimodal performance with only 4 billion parameters but also easily transfers to downstream tasks across various domains at low marginal cost.

(2) We develop several design features for Mini-InternVL, including a lightweight vision encoder---InternViT-300M, that is robust for various visual domains. Additionally, we introduce a simple but effective paradigm that standardizes model architecture, data format, and training schedule for effective downstream task transfer.

(3) We comprehensively evaluate our models through extensive experiments on general and domain-specific benchmarks. These results show that our multimodal models achieve 90\% of the performance using significantly fewer parameters on general multimodal benchmarks. For specific domain tasks, with minimal computational cost for fine-tuning, they can rival closed-source commercial models. We conduct a series of ablation studies to explore the impact of data sample size on domain adaptation, hoping to provide insights into the application of MLLMs in specialized domains.

\section{Related Works}

\paragraph{Multimodal Large Language Models.}

Benefiting from the advancement of LLMs, the MLLMs have also achieved great progress.
Early works~\cite{liu2023interngpt,shen2024hugginggpt,wu2023visualgpt} consider multi-modal understanding as one of the tool usage tasks and prompt the LLMs to ask other models to write a caption about the corresponding input modality so that LLMs could understand the multi-modal input.
To effectively utilize the ability of pre-trained LLMs and VFMs, a series of works~\cite{li2022blip,li2023blip2,liu2023llava,Qwen-VL,chen2023internvl,peng2023kosmos2,wang2024mmniah} propose to use a connector to align the embedding space between them, which achieve promising performance under a controllable cost.
Another series of work~\cite{alayrac2022flamingo,wang2023cogvlm,tian2024mminterleaved,llama3modelcard} extend pre-trained LLMs with extra layers to fuse the vision features, which reduce the number of required visual tokens inputted into LLMs while introducing extra training cost.

Recently, some works, such as Fuyu~\cite{fuyu-8b}, MoMa~\cite{lin2024moma} and Chameleon~\cite{team2024chameleon}, propose a vision encoder-free architecture. This type of architecture consists of a single transformer model, which is used to process both visual and textual information simultaneously without requiring an additional encoder, making it more deployment-friendly.
Despite these advancements, the heavy inference cost of these MLLMs hinders their application in downstream tasks.
To address such issue, a series of lightweight MLLMs, such as MiniCPM-V~\cite{yao2024minicpm}, are proposed.
However, since most of them use CLIP-L~\cite{radford2021learning} as the vision encoder, which is only trained on the natural image domain, these models are limited to the general domain and fail to generalize to other domains.
In this work, we propose InternViT-300M, which is distilled from InternViT-6B and trained on a diverse set of image domains.

\paragraph{Vision Foundation Models for MLLMs.}
From a vision-centric perspective, most MLLMs utilize vision models such as CLIP~\cite{radford2021learning, EVA-CLIP} and SigLIP~\cite{zhai2023sigmoid}, which are trained on large-scale web image-text data. However, such vision encoders face significant limitations in terms of parameter scale and representational ability. 
Several studies have explored this issue. For instance, Tong \emph{et al.}~\cite{tong2024eyes} identified significant differences in the visual patterns of CLIP and DINOv2~\cite{oquab2023dinov2}, leading to the development of a mixture-of-features module that integrates these two VFMs. LLaVA-HR~\cite{luo2024feast} introduces a dual-branch vision encoder that employs CLIP-ViT for low-resolution pathways and CLIP-ConvNext for high-resolution pathways. 
Similarly, DeepSeek-VL~\cite{lu2024deepseekvl} utilizes a dual vision encoder design, incorporating SigLIP-L for low-resolution images and SAM-B~\cite{kirillov2023segany} for high-resolution images. 

However, these methods involve excessively complex pathways, which complicates the practical application of the models. Moreover, such approaches do not resolve the issue of vision encoders lacking comprehensive visual knowledge across various domains. 
In contrast, InternViT~\cite{chen2023internvl} implements progressive image-text alignment, and acquires representational capabilities across multiple domains by performing generative training on datasets spanning various fields. We propose injecting visual knowledge from the capable vision encoder into lightweight visual models, thus avoiding the computational expense associated with iterative generative pre-training.

\paragraph{Domain-Specialized Adaptation of MLLMs.}
Several methods have been explored to apply MLLMs to specific domains, such as GeoChat~\cite{kuckreja2023geochat} and EarthGPT~\cite{zhang2024earthgpt} for remote sensing, LLaVA-Med~\cite{li2023llavamed} and Qilin-Med-VL~\cite{Liu2023QilinMedVLTC} for the medical images, ChemVLM~\cite{li2024chemvlm} for chemistry, DriveVLM~\cite{tian2024drivevlm}, DriveMLM~\cite{wang2023drivemlm} and DriveGPT4~\cite{xu2023drivegpt4} for autonomous driving. 
Although these methods have achieved promising results, they involve modifications to model architectures, the collection of extensive domain-specific training data, or customization of the training process for the target domain. Nonetheless, there is still no universally accepted framework for the downstream adaptation of MLLMs. we propose a straightforward yet effective transfer learning paradigm, aiming to prevent significant disparities among MLLMs in different fields that hinder interoperability.

\begin{figure*}[t]
  \centering
    \includegraphics[width=\textwidth,trim={0cm 5cm 0cm 3cm}, clip]{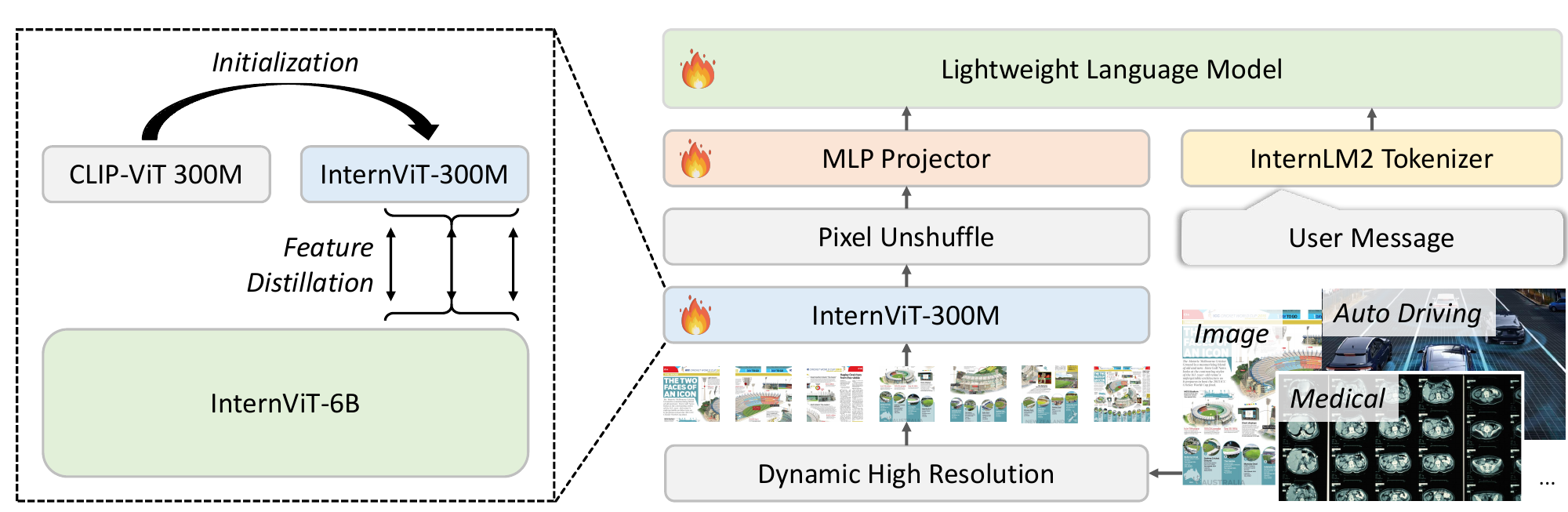}
  \caption{\textbf{Training method and architecture of \miniinternvl.} Left: We employ InternViT-6B~\cite{chen2023internvl} as the teacher model to perform knowledge distillation on the student model. Right: \miniinternvl adopts the ViT-MLP-LLM architecture similar to popular MLLMs~\cite{yao2024minicpm,lu2024deepseekvl, liu2023llava,Qwen-VL,wang2023allseeing,wang2024allseeingv2}, combining InternViT-300M with a series of lightweight LLMs through an MLP projector. Here, we employ a simple pixel unshuffle to reduce the number of visual tokens to one-quarter}

  \label{fig:architecture}

\end{figure*}

\section{Method}

In this section, we introduce \miniinternvl, a series of lightweight multimodal large language models (MLLMs).
\secref{sec: architecture} provides a comprehensive overview of \miniinternvl. Then, \secref{sec: injection} details InternViT-300M, a lightweight vision model developed through knowledge distillation, which inherits the strengths of a powerful vision encoder. Finally, \secref{sec: framework} describes a transfer learning framework designed to enhance the model's adaptation to downstream tasks.

\subsection{\miniinternvl}
\label{sec: architecture}

As shown in \figref{fig:architecture}, \miniinternvl consists of three main components: visual encoder, MLP projector, and LLM.
We employ InternViT-300M as our visual encoder, a lightweight vision model that inherits the capabilities of a powerful vision encoder. Based on InternViT-300M, we develop three versions of \miniinternvl: \miniinternvl-1B, \miniinternvl-2B, and \miniinternvl-4B, which are respectively connected to the pre-trained Qwen2-0.5B~\cite{qwen2}, InternLM2-1.8B~\cite{cai2024internlm2}, and Phi-3-mini~\cite{abdin2024phi}. Similar to other open-source MLLMs~\cite{liu2023llava,chen2024far,li2024llava,wang2024allseeingv2}, \miniinternvl employs an MLP projector to connect the vision encoder and the LLMs.

We adopt a dynamic resolution input strategy similar to that of InternVL 1.5~\cite{chen2024far}, which improves the model's ability to capture fine-grained details. 
We also apply a pixel unshuffle operation to reduce the number of visual tokens to one-quarter of the original. Consequently, in our model, a 448$\times$448 image is represented by 256 visual tokens, enabling it to process up to 40 image tiles (\ie, 4K resolution).

The training of \miniinternvl consists of two stages: (1) Language-image alignment: We keep only the MLP component unfrozen during this stage. Following InternVL 1.5~\cite{chen2024far}, we use a diverse range of training datasets that encompass various tasks, including captioning, detection, grounding, and OCR. The diversity of these datasets ensures robust pre-training of \miniinternvl, enabling the model to handle a variety of linguistic and visual elements across different tasks. 
(2) Visual instruction tuning: We carefully select datasets to enhance the model's performance across a broad spectrum of multimodal tasks, similar to InternVL 1.5. These tasks include image captioning, chart interpretation, OCR, and cross-disciplinary reasoning. We conduct full-parameter fine-tuning with these datasets, further injecting world knowledge and teaching models to follow user instructions.

\subsection{InternViT-300M}

\label{sec: injection}

\begin{table}[t]
    \centering
    \caption{Datasets used in knowledge distillation of vision encoder.}
    \vspace{0.5em}
    \label{tab:distil_dataset}
      \scriptsize
      \setlength{\tabcolsep}{2em}
    \begin{tabular}{l|l}
       \toprule
         Type & Dataset \\
         \midrule
         \multirow{4}{*}{Natural images} & Laion~\cite{schuhmann2022laion5b}, COYO \cite{byeon2022coyo}, GRIT~\cite{peng2023kosmos2}, COCO~\cite{chen2015cococaption},\\
         & LVIS~\cite{gupta2019lvis}, Objects365~\cite{shao2019objects365}, \\ 
         & Flickr30K~\cite{plummer2015flickr30k}, VG~\cite{krishna2017visual}, All-Seeing~\cite{wang2023allseeing,wang2024allseeingv2}, \\
         & MMInstruct~\cite{liu2024mminstruct}, LRV-Instruction~\cite{liu2023aligning} \\     
         \midrule
         \multirow{7}{*}{OCR} & TextCaps~\cite{sidorov2020textcaps}, Wukong-OCR~\cite{gu2022wukong}, CTW~\cite{yuan2019ctw},\\
          & MMC-Inst~\cite{liu2023mmcinst}, LSVT~\cite{sun2019lsvt}, ST-VQA~\cite{biten2019stvqa}, \\
         & RCTW-17~\cite{shi2017rctw17}, ReCTs~\cite{zhang2019rects}, ArT~\cite{chng2019art}, \\
         & SynthDoG~\cite{kim2022synthdog}, LaionCOCO-OCR~\cite{schuhmann2022laioncoco}, \\
         & COCO-Text~\cite{veit2016cocotext}, DocVQA~\cite{mathew2021docvqa}, TextOCR~\cite{singh2021textocr},  \\
         & LLaVAR~\cite{zhang2023llavar}, TQA~\cite{kembhavi2017you}, SynthText~\cite{Gupta16}\\
         & DocReason25K~\cite{hu2024mplug}, Common Crawl PDF\\

         \midrule
         \multirow{3}{*}{Chart}& AI2D~\cite{kembhavi2016ai2d}, PlotQA~\cite{methani2020plotqa}, InfoVQA~\cite{mathew2022infographicvqa}, \\
         &ChartQA~\cite{masry2022chartqa}, MapQA~\cite{chang2022mapqa}, FigureQA~\cite{kahou2017figureqa}, \\
         &IconQA~\cite{lu2021iconqa}, MMC-Instruction~\cite{liu2023mmc} \\
         
         \midrule
         \multirow{4}{*}{Multidisciplinary} & CLEVR-Math/Super~\cite{li2023superclevr,lindstrom2022clevrmath}, GeoQA+~\cite{cao2022augmented}, \\
          & UniChart~\cite{masry2023unichart}, ScienceQA~\cite{lu2022scienceqa}, Inter-GPS~\cite{lu2021inter},\\
          & UniGeo~\cite{chen2022unigeo}, PMC-VQA~\cite{zhang2023pmc}, TabMWP~\cite{lu2023dynamic},
            \\
            & MetaMathQA~\cite{yu2023metamath}\\
         \midrule
         \multirow{2}{*}{Other} & Stanford40~\cite{yao2011human}, GQA~\cite{hudson2019gqa}, MovieNet~\cite{huang2020movienet}, \\ 
         & KonIQ-10K~\cite{koniq10k}, ART500K~\cite{mao2017deepart}, ViQuAE~\cite{lerner2022viquae}\\
         \bottomrule 
    \end{tabular}

\end{table}

Most existing MLLMs~\cite{yao2024minicpm,lu2024deepseekvl, liu2023llava,Qwen-VL,wang2023allseeing,wang2024allseeingv2} employ vision encoders that are trained on web-scale image-text paired data, such as CLIP, to obtain their representations. These encoders lack comprehensive knowledge of the visual world, which needs to be acquired through iterative generative pre-training in conjunction with LLMs.
Unlike other approaches that enhance the visual foundation models by using auxiliary pathways~\cite{tong2024eyes, lu2024deepseekvl,li2024mgm}, our method directly leverages a powerful vision model that has undergone generative training on diverse datasets to transfer knowledge to a lightweight vision model. Specifically, we use InternViT-6B as the teacher model and initialize the student model's weights using {CLIP-ViT-L-336px}. We align the representations of the student model with those of the teacher model by computing the negative cosine similarity loss between the hidden states of the last $K$ transformer layers. The resulting model is named InternViT-300M.

The primary goal of this knowledge transfer is to inherit the pre-training knowledge embedded in InternViT-6B. To achieve this, we curate a dataset sourced from a diverse range of publicly accessible resources, as detailed in \tabref{tab:distil_dataset}. This dataset comprises four main types of data: natural images, OCR images, charts, and multi-disciplinary images. All images are resized to a resolution of 448
$\times$448, and dynamic resolution~\cite{chen2024far} is disabled for training efficiency. Ultimately, we develop a vision encoder, termed InternViT-300M, which is infused with diverse knowledge and is adaptable to various language models.

\subsection{Domain Adaptation}

\label{sec: framework}
\begin{figure*}[t]
  \centering
    \includegraphics[width=\textwidth, trim={0.5cm 2cm 0.5cm 1cm},clip]{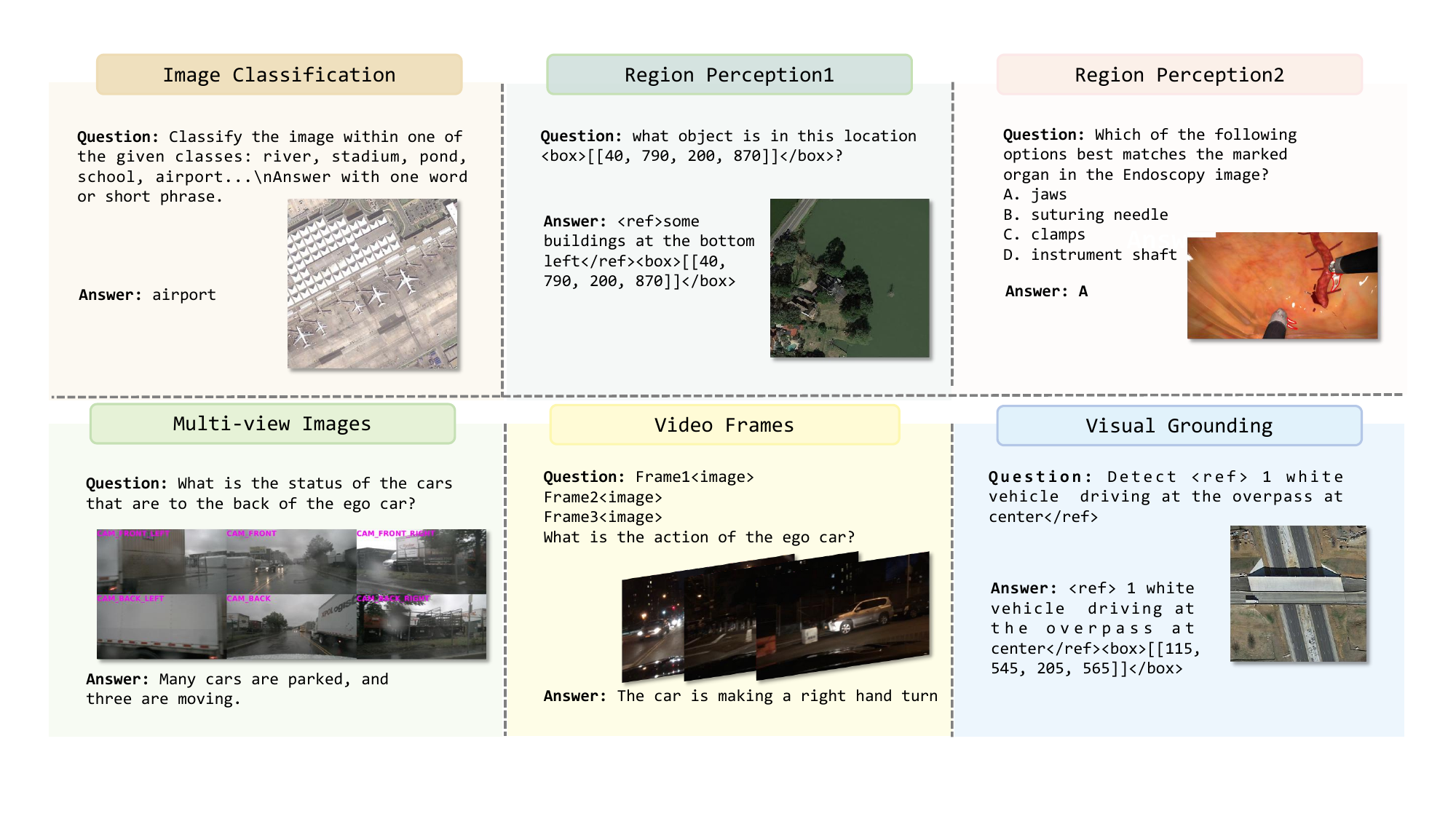}
  \caption{\textbf{The data format of our adaptation framework.} We formulate other visual tasks (\ie, image classification, region perception, multi-view image tasks, video-related tasks, and visual grounding) into VQA format in our framework.}
  \label{fig: framework}
\end{figure*}

Although many studies~\cite{kuckreja2023geochat,zhang2024earthgpt,li2023llavamed,xu2023drivegpt4} have successfully applied MLLMs to downstream tasks, a universally accepted framework for adapting MLLMs to these applications has yet to be established.
Differences in model design, data formats, and training strategies across various domains result in significant heterogeneity among MLLMs, making standardization challenging. To address this issue, we propose a straightforward yet effective transfer learning framework.

\paragraph{Data Format.} Instruction tuning is a crucial training stage to teach models to follow user instructions, where the training data is formulated as visual question answering (VQA) and conversation format. VQA datasets of the downstream tasks, such as RSVQA~\cite{lobry2020rsvqa} and PMC-VQA~\cite{zhang2023pmc}, are directly utilized as instruction-following data. For other conventional tasks, as shown
in \figref{fig: framework}, we formulate them into VQA format according to the following approaches separately:

\textbf{(1) Image Classification Tasks.} In most traditional classification tasks within specialized domains, a wide range of technical terms are involved. In the majority of cases, we can easily format the classification task as a multiple-choice question. Given an image \texttt{<image>}, the set of candidate labels $O$, and the ground truth $G \in O$, the template can be expressed as:
\begin{center}
    \makebox[0pt][c]{%
    \parbox{0.9\textwidth}{%
    \textbf{USER}: \texttt{[Image][Prompt\_Prefix][Candidate Labels][Prompt\_Suffix]} \\
    \textbf{ASSISTANT}: \texttt{[Ground Truth]}
    }
    }
\end{center}

A direct example can be seen in our approach to remote sensing image classification, where we utilize prompts such as ``\texttt{Classify the image within one of the given classes: dense residential area, ..., school. Answer with one word or short phrase.}'', as shown in \figref{fig: framework}. This method transforms image classification tasks into multiple-choice questions. For behavior prediction of the ego vehicle in autonomous driving data, we draw inspiration from DriveLM~\cite{sima2023drivelm} by employing templates like ``\texttt{Predict the behavior of the ego vehicle. Please select the correct answer from the following options: A. The ego vehicle is going straight. The ego vehicle is not moving. B. ...}''.

\textbf{(2) Visual Grounding Tasks.}
The native support for the visual grounding task in \miniinternvl
allows the use of a special token, \texttt{<ref></ref>}, to enclose the name of the object to be detected. With this token, the model can be directed to provide the object's location enclosed within \texttt{<box></box>} in the format [[\texttt{x1}, \texttt{y1}, \texttt{x2}, \texttt{y2}]], where the coordinates range from 0 to 1000. This approach enables us to convert object grounding and referring expression detection into a conversational format.
We extensively apply this format to remote sensing instruction data. For example, for the referring expression ``\texttt{1 overpass near some trees at the center}'', we use ``\texttt{Detect <ref>1 overpass near some trees at the center</ref>}'' as the instruction and ``\texttt{<ref>1 overpass near some trees at the center</ref><box>[[x1, y1, x2, y2]]</box>}'' as the response.

\textbf{(3) Region Perception Tasks.} Region-level conversation tasks are prevalent in specialized domains. These tasks involve supplying the model with spatial location information, in addition to the question input. The model is required to focus on objects within the specified attention region to generate a response. 
Specifically, there are two implementation methods. The first method involves directly annotating the location on the image using bounding boxes, masks, or contours, as illustrated in the ``Region Perception2'' of \figref{fig: framework}. The second method denotes the object within the question by \texttt{<box>[[x1,y1,x2,y2]]</box>}, where the coordinates are normalized between 0 and 1000. This notation guides the model's attention to specific regions within the image, enabling it to perform tasks such as region-level captioning and region-specific VQA.

For instance, in remote sensing applications, the goal is to train the model to identify objects within specific coordinates \texttt{[x1, y1, x2, y2]}. To achieve this, we use a prompt such as ``\texttt{What object is in this location<box>[[x1, y1, x2, y2]]</box>}'' as the input instruction, with ``\texttt{<ref>object name</ref><box>[[x1, y1, x2, y2]]</box>}'' serving as the label.

\begin{figure*}[t]
  \centering
    \includegraphics[width=\textwidth, trim={2.1cm 4.1cm 2cm 3.3cm},clip]{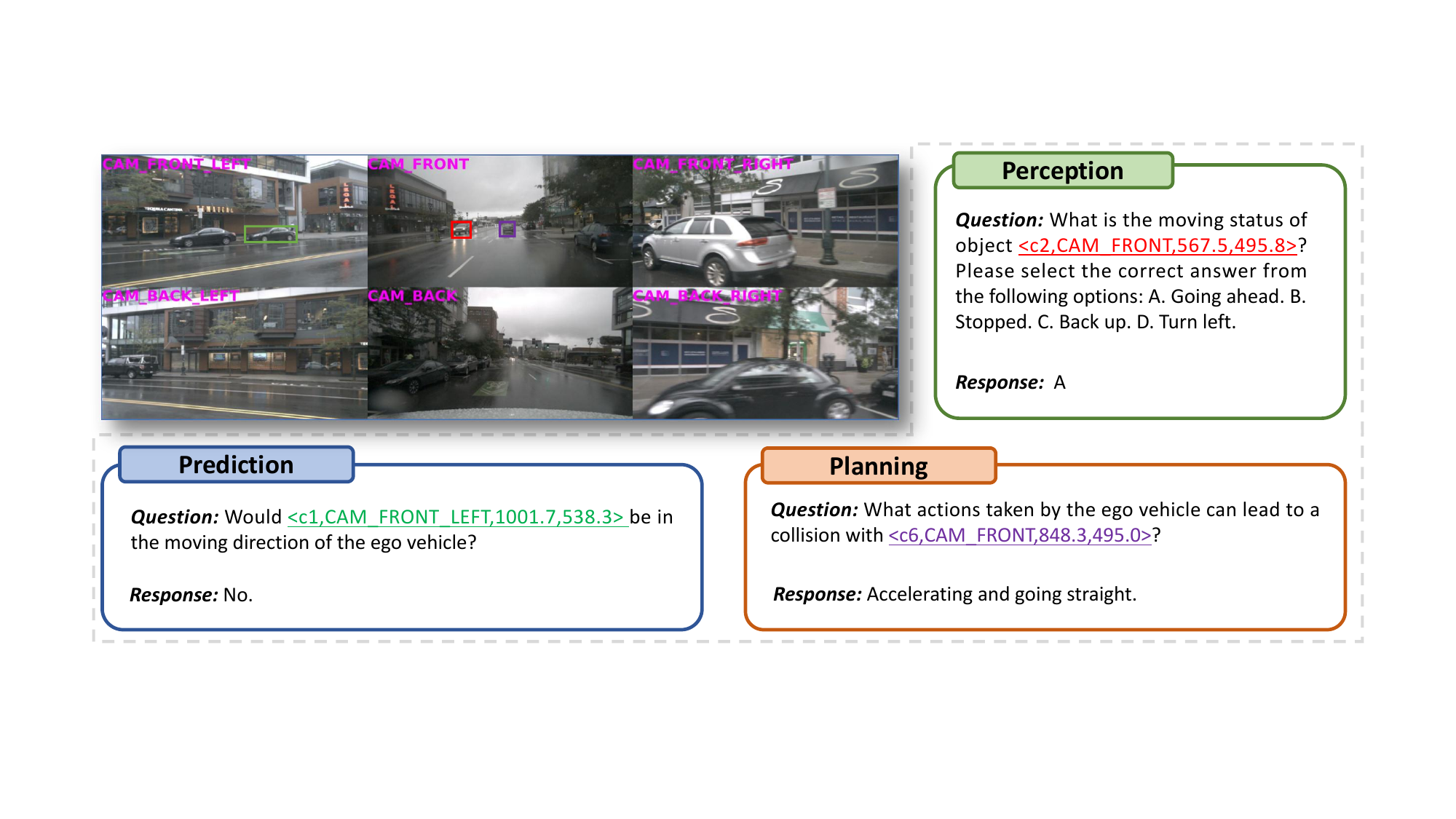}
  \caption{\textbf{Qualitative Results of \internvlad.} In the upper left corner is a multi-view image from DriveLM-nuScenes version-1.1~\cite{sima2023drivelm} after data processing. The color of the bounding boxes corresponds to the font color of the \textbf{c tags} for the objects in question (note that the input images do not contain these manually drawn bounding boxes). We show the predicted answers of our model in perception, prediction, and planning tasks. The model's outputs align with human driving behavior. }
  \label{fig: drivelm}
\end{figure*}

\textbf{(4) Multi-View Images.} In autonomous driving, the images are captured from six different viewpoints. As shown in \figref{fig: drivelm}, we effectively utilize 
dynamic resolution to accommodate this type of data. Specifically, \miniinternvl supports splitting images into 448$\times$448-sized tiles based on their aspect ratio. Consequently, we resize each image to 896$\times$448 pixels and then, as illustrated, combine these images in a fixed sequence, resulting in a final resolution of 2688$\times$896. This means that the images are automatically processed into 12 tiles, and an additional thumbnail is added to provide the model with global context. Furthermore, we label each viewpoint image with text indicating its camera position, such as ``CAM\_FRON''.

\textbf{(5) Video Frames.} \miniinternvl supports video frames in an interleaved image format. We represent the frame sequence using a template such as ``\texttt{Frame1: <img><IMG\_CONTEXT></img> Frame2: <img><IMG\_CONTEXT></img>.}'', where \texttt{<IMG\_CONTEXT>} denotes the image tokens. For each image at a resolution of 448$\times$448, the model can accommodate sequences of up to 40 frames.

\paragraph{Training Strategy.}

During the domain adaptation phase, we perform full-parameter fine-tuning on \miniinternvl. For a domain-specific application scenario, we convert corresponding data into the required format and incorporate it into our training dataset.
Adding a certain proportion of general multimodal data during the domain adaptation phase will not affect the performance in the specific domain, while retaining the model's general multimodal capability.
In our experiments, we find that adding general data can improve the generalization ability of the model on other tasks. Therefore, when performing domain adaptation, we can choose the appropriate general data ratio on the premise of balancing computational overhead and performance.

\begin{table}[!t]

\centering
\caption{\textbf{Comparison with other models on multimodal benchmarks.} We evaluate models using the InternVL and VLMEvalKit~\cite{duan2024vlmevalkit} repositories. AI2D~\cite{kembhavi2016ai2d}, ChartQA~\cite{masry2022chartqa}, DocVQA~\cite{mathew2021docvqa}, InfoVQA~\cite{mathew2022infographicvqa}, and MMBench~\cite{liu2023mmbench} are tested with InternVL, while MathVista~\cite{lu2023mathvista} and OCRBench~\cite{liu2023ocrbench} use VLMEvalKit. For MMMU~\cite{yue2023mmmu}, we report scores from the OpenCompass leaderboard. The Avg. Score is the average of the scores from all tested benchmarks, with the OCRBench score divided by 10. The values in parentheses represent the relative parameters and performance of \miniinternvl compared to {InternVL2-Llama3-76B~\cite{chen2024far}}, which is considered as 100\%.}
\vspace{0.5em}
\label{tab:main_results}
 \scriptsize
\setlength{\tabcolsep}{1.3pt}
\begin{tabular}{l|c|c|ccccccccc|c}
\toprule
\multirow{2}{*}{model} & open-& \multirow{2}{*}{\#param} & MMMU & MathVista & \multirow{2}{*}{AI2D}&\multirow{2}{*}{ChartQA}        &   \multirow{2}{*}{DocVQA}   & \multirow{2}{*}{InfoVQA}     & OCR-      &   MMB- &MMB-     & \multirow{2}{*}{Avg. Score} \\
&\strut  source  &  & (val)  &  (testmini)  &   &     &   &  & Bench  & EN  & CN& \\
\midrule
GPT-4V-0409~\cite{openai2023gpt4}&\no &-&  61.7 & 58.1 & 89.4 & 78.1 & 87.2 & - & 678 & 81.0 & 80.2 & 75.4 \\
Gemini-Pro-1.5~\cite{reid2024gemini1_5} &\no  &-& 60.6 & 57.7 & 80.3 & 81.3 & 86.5 & 72.7 & 754 & 73.9 & 73.8 & 73.6\\
Claude3.5-Sonnet~\cite{claude3_5series2024} & \no &-&65.9 & 67.7 & 94.7 & 90.8 & 95.2 & - & 788 & 79.7 & 80.7 & 81.7 \\
GPT-4o~\cite{openai2024gpt4o} &\no &-& 69.2 & 63.8 & 94.2 & 85.7 & 92.8 & - & 736 & 83.4 & 82.1 & 80.6 \\
Cambrian-1~\cite{tong2024cambrian1} &\yes &-& 50.4 & 53.2 & 79.7 & 75.6 & 75.5 & - & 600 & 81.4  & - & 68.0\\
DeepSeek-VL-1.3B~\cite{lu2024deepseekvl} &\yes &2B & 33.8 & 29.8 &51.5&-&-&-&413&64.6 &62.9&47.3\\
MiniCPM-V 2.0~\cite{yao2024minicpm} &\yes&3B& 38.2&38.7 &62.9&-&71.9&-&605&69.1 & 66.5 &58.3 \\
Qwen2-VL-2B~\cite{Qwen2VL}& \yes &2B& 42.2&43.0& 74.7 &73.5 &90.1&65.5&794&74.9 & 73.5& 68.5 \\
\midrule

InternVL2-Llama3-76B~\cite{chen2024far}
&\yes &76B &58.2  & 65.5  & 87.6 & 88.4  & 94.1  & 82.0  & 839  & 86.5 & 86.3  & 81.4\\

\miniinternvl-1B & \yes&1B (1\%)&36.7 & 37.7 & 64.1 & 72.9  & 81.7  & 50.9 & 754  & 65.4 & 60.7  & 60.6 (74\%) \\
\miniinternvl-2B & \yes&2B (3\%)&36.3& 46.3  & 74.1 & 76.2 & 86.9 & 58.9 & 784 & 73.2 & 70.9 & 66.8 (82\%)\\
\miniinternvl-4B & \yes&4B (5\%)&48.3 & 58.6  & 78.9 & 81.5  & 89.2 & 67.0 & 788  & 78.6 & 73.9 & 72.8 (90\%)\\
\bottomrule

\end{tabular}

\end{table}

\section{Experiments}

\label{sec:experiments}

In this section, we begin by conducting a comprehensive comparison of our \miniinternvl with leading multi-modal large language models (MLLMs) on representative vision-language benchmarks (\secref{sec:main_results}). Following this, in \secref{sec:transfer}, we apply the domain adaptation framework introduced in \secref{sec: framework} to transfer our models to three specialized domains: autonomous driving (\secref{sec:Autonomous driving} and \secref{sec:temporal}), medical images (\secref{sec:medical}), and remote sensing (\secref{sec: Remote}). Additionally, we perform an extensive ablation study to explore the impact of data sample size and model size on domain adaptation (\secref{sec:ablation}).

\subsection{Results on General Multimodal Benchmark.}
\label{sec:main_results}

\paragraph{Settings.}
In this section, we present a comprehensive evaluation of our model's multimodal understanding and reasoning capabilities across a variety of benchmarks. The benchmarks used in our study are categorized into four distinct types: \textbf{OCR-related tasks}, including DocVQA~\cite{mathew2021docvqa}, OCRBench~\cite{liu2023ocrbench} and InfographicVQA~\cite{mathew2022infographicvqa}; \textbf{chart and diagram understanding}, including AI2D~\cite{kembhavi2016ai2d} and ChartQA~\cite{masry2022chartqa}; \textbf{general multimodal tasks}, such as MMBench~\cite{liu2023mmbench}; and \textbf{multimodal reasoning}, including MMMU~\cite{yue2023mmmu} and MathVista~\cite{lu2023mathvista}. Additionally, we calculate the average score across these benchmarks, with the OCRBench score divided by 10.

\paragraph{Results.}
As shown in \tabref{tab:main_results}, \miniinternvl demonstrates strong performance across the majority of benchmarks.
Our smallest model contains only 1 billion parameters, yet it demonstrates performance comparable to 2 billion parameter models, such as DeepSeek-VL-1.3B and MiniCPM-V 2.0. Compared to other lightweight models, our \miniinternvl-4B excels across most benchmarks, particularly in MMbench, ChartQA, DocVQA, and MathVista, where its performance is on par with commercial models like Gemini-Pro-1.5.
Notably, compared to InternVL2-Llama3-76B, which utilizes the larger InternViT-6B, \miniinternvl achieves approximately 90\% of its performance while using 5\% parameters. This highlights the effectiveness of our knowledge distillation strategy.

\subsection{Transfer to Various Specialized Domains}
\label{sec:transfer}

\subsubsection{Multi-View Image-Based Autonomous Driving}
\label{sec:Autonomous driving}

\begin{table}[t]
  \centering
  \small
  \caption{\textbf{The sources of our general datasets}. In each task, we sample a specific amount of data from the data source at a predetermined ratio to balance the training dataset. 
  In each domain adaptation case, we sample a certain amount of data from each data source.}
  \vspace{0.5em}
    \setlength{\tabcolsep}{6pt}
  \begin{tabular}{c|cc}
    \toprule
     Data Source  & Size & Data Type \\
    \midrule
    ShareGPT4V~\cite{chen2023sharegpt4v} & 767K & Captioning \\
    AllSeeingV2~\cite{wang2024allseeingv2} & 127K & Grounding VQA \\
    LLaVA-Instruct-ZH\protect \footnotemark & 158K & VQA \\
    DVQA~\cite{kafle2018dvqa} & 200K & Diagram VQA \\
    ChartQA~\cite{masry-etal-2022-chartqa} & 18K & Diagram VQA \\
    AI2D~\cite{kembhavi2016diagram} & 12K & Diagram VQA  \\
    DocVQA~\cite{mathew2021docvqa} & 10K & Document VQA  \\
    GeoQA+~\cite{cao-xiao-2022-augmented} & 72K & Geometric VQA\\
    SynthDoG-EN~\cite{kim2022donut} & 30K & OCR \\
     \bottomrule
  \end{tabular}

   \label{tab: general_data}
\end{table}

\footnotetext{\url{https://huggingface.co/datasets/openbmb/llava_zh}}

\begin{table}[!t]

  \centering
  \caption{\textbf{The results on driving with language official leaderboard~\cite{sima2023drivelm}.} ``DA'' means model after domain adaptation on DriveLM.
  The other results in the table are taken from the CVPR 2024 Autonomous Driving Challenge leaderboard. 
  MTMM\textsuperscript{$\dagger$}, MMFM\_AD, and Team NVIDIA are team names on the challenge leaderboard, which we use to represent their methods.} 
  \vspace{0.5em}
   \scriptsize
    \setlength{\tabcolsep}{3pt}
  \begin{tabular}{c|c|cccccccccc}
   \toprule      
    Method     & \#Param & Accuracy   & ChatGPT  & Bleu 1  & Bleu 2  & Bleu 3  & Bleu 4  & ROUGE L & CIDEr & Match     & Final Score  \\
    \midrule 
    InternVL4Drive-v2~\cite{li2024driving} &26B& 0.7339 & 65.25 & 0.7787 &0.7176& 0.6608 &0.6059 &0.7449 &0.2061 &47.65 &0.6002 \\
    MTMM\textsuperscript{$\dagger$} &-& 0.7473&  65.59 & 0.76 &0.70 & 0.64 & 0.59 & 0.74 & 0.18 & 0.45 & 0.5974 \\
    Team NVIDIA &-& 0.7746&  59.89 & - &-& - & - &- & - &- & 	0.5884 \\
    MMFM\_AD	&-& 	0.6658&  63.92 & - &-& - & - &- & - &- & 	0.5732 \\
    \midrule
    \miniinternvl-4B &4B&0.0&54.45&0.2405&0.0801&0.0252&0.0084&0.1927&0.0018&34.30&0.3051\\
    InternVL2-Llama3-76B &76B & 0.0 &52.50&0.2100&0.0884&0.0249&0.0078&0.1848&0.0001& 34.22 &0.2963\\

    \internvlad-1B & 1B &0.7007  & 63.84  & 0.7362 & 0.6767 & 0.6214 & 0.5678 & 0.7365 & 0.1669 & 39.76  & 0.5686 \\
    \internvlad-2B & 2B & 0.7628  & 65.23  & 0.7616 & 0.7012 & 0.6452 & 0.5908 & 0.7447 & 0.1914 & 43.24  & 0.5958 \\
    \internvlad-4B & 4B &0.7296  & 63.97  & 0.7642 & 0.7032 & 0.6463 & 0.5914 & 0.7427 & 0.1976 & 42.16  & 0.5821 \\
    \bottomrule
  \end{tabular}

    \label{tab:results_on_drivelm} 

\end{table}
\paragraph{Settings.}

We select DriveLM-nuScenes version 1.1~\cite{sima2023drivelm} as our training dataset, which contains 317K training samples and encompasses various aspects of the driving process. This dataset includes data for perception, prediction, and planning, offering a comprehensive understanding of autonomous driving scenarios. 

In DriveLM-nuScenes, the images are captured from six different viewpoints. We effectively utilize dynamic resolution features to accommodate this type of data. Specifically, \miniinternvl supports splitting images into 448$\times$448-sized tiles based on their aspect ratio. As illustrated in \figref{fig: drivelm}, we resize the image of each view to 896$\times$448 pixels and then combine these images in a fixed sequence, resulting in a final resolution of 2688$\times$896.
This means that the images are automatically processed into 12 tiles, and an additional thumbnail is added to provide the model with global context. Furthermore, we mark the image of each view with text indicating its camera position, such as ``CAM\_FRON''.

As shown in \figref{fig: drivelm}, DriveLM-nuScenes contains QA pairs with coordinates, thus we need to normalize them to a range of 0 to 1000 to align with the output of \miniinternvl.
In the dataset, objects are represented by \textbf{c tags}. We use a tailored prompt: ``\texttt{Objects are encoded using <c, CAM, [cx,cy]>, where c is the identifier, CAM indicates the camera where the object's center point is situated, and x, y  represent the horizontal and vertical coordinates of the center point of the 2D bounding box.}'' to guide the model on the composition of \textbf{c tags}. The ground truth responses typically include bounding box annotations for questions requiring a list of all objects. Therefore, We annotate the bounding box as \texttt{<box>[[x1,y1,y2,y3]]</box>}, where \texttt{<box>} and \texttt{</box>} are special tokens in \miniinternvl. Additionally,  we incorporate general datasets into the training set, to prevent the model from losing its general domain perception capabilities, maintaining a 1:4 ratio of general to domain-specific data. The sources of the general datasets are shown in \tabref{tab: general_data}.

Finally, we conduct full-parameter fine-tuning of \miniinternvl using 8 A100 GPUs, training the model for 1 epoch with a learning rate of 1e-5.
We report the performance of our model after transfer learning on the CVPR 2024 Autonomous Driving Challenge~\cite{sima2023drivelm}. Furthermore, we evaluate our model on autonomous driving scenarios from MME-Realworld~\cite{zhang2024mme}, where we separately assess its performance on Perception and Reasoning tasks.

\begin{wraptable}{r}{0.55\textwidth}
\small
\centering
\vspace{-1.3em}
 \caption{\textbf{Results on Autonomous Driving domain of MME-RealWorld.} ``DA'' means model after domain adaptation on DriveLM.}
\setlength{\tabcolsep}{1.2pt}
  \begin{tabular}{c|cc|c}
     \toprule     
    Method & Perception  & Reasoning & Avg. \\
      \midrule 
    GPT-4o~\cite{openai2024gpt4o} &21.14 &26.41 & 24.60 \\
    Claude 3.5 Sonnet~\cite{claude3series2024} &32.43&31.92& 32.10 \\
    \midrule 
    LLaVA-OneVision-7B~\cite{li2024llava}& 45.77 & 34.08& 41.75 \\
    Qwen2-VL-7B~\cite{Qwen2VL} &34.62&31.47& 33.54\\
    InternVL2-Llama3-76B~\cite{chen2024far} & 47.46& 35.71 & 44.30 \\
    \midrule 
    \miniinternvl-1B &31.34&22.47&28.96 \\
    \miniinternvl-2B &39.86	&30.13	&37.25 \\
    \miniinternvl-4B &38.96	&33.11	&37.39 \\

    \internvlad-1B &42.95	&27.75	&38.87 \\

    \internvlad-2B &50.74	&40.48	&47.98 \\ 

    \internvlad-4B &53.14	&39.14	&49.38 \\

    \bottomrule
  \end{tabular}

\label{tab: mme-rw}  
\end{wraptable}

\paragraph{Results.}

We test our model using DriveLM-nuScenes-version-1.1-val \cite{sima2023drivelm}, and the results are presented in \tabref{tab:results_on_drivelm}.
Our final score of \miniinternvl-2B is 0.5958, which is comparable to the best result on the CVPR 2024 Autonomous Driving Challenge Leaderboard\footnote{\url{https://opendrivelab.com/challenge2024/\#driving_with_language}}, InternVL4Drive-v2~\cite{li2024driving}. Notably, our model uses only one-tenth of the parameters of InternVL4Drive-v2.

Our model scores slightly lower in the Match metric, which might be due to \miniinternvl's lack of proficiency in predicting object center points. InternVL4Drive-v2~\cite{li2024driving} offers a viable solution by using Segment Anything~\cite{kirillov2023segment} to convert object center points into object bounding boxes. In \figref{fig: drivelm}, we show the predicted answers of our model in perception, prediction, and planning tasks, demonstrating alignment with human driving behavior.
Furthermore, our 4B-parameter model performs similarly to our 2B-parameter model. We attribute this potentially to the limitations of the existing training data and evaluation criteria, which might constrain larger models from achieving significant performance gains.

As shown in \tabref{tab: mme-rw}, in the autonomous driving scenarios of MME-Realworld, we observe that even when using only DriveLM as domain-specific training data, our model achieves an improvement of over 10 points. The transferred model surpasses the best-performing model on this task, LLaVA-OneVision-7B~\cite{li2024llava}, as well as several commercial closed-source models such as GPT-4o~\cite{openai2024gpt4o} and Claude 3.5 Sonnet~\cite{claude3series2024}, demonstrating the strong generalization capability of our model.

\subsubsection{Autonomous Driving with Temporal Information}
\label{sec:temporal}
\paragraph{Settings.}

Using single-frame images alone is insufficient for accurate perception and prediction of vehicle behavior. Therefore, we explore temporal expansion. Specifically, we utilize instruction-following data constructed by DriveGPT4~\cite{xu2023drivegpt4} from the BDD-X dataset~\cite{kim2018textual} as our training set, which comprises 26K video clips. Multiple video frames are organized as described in \secref{sec: framework}. Each training sample contains four aspects of question-answer data: Action Description, Action Justification, Speed Signal Prediction, and Turning Angle Signal Prediction. 
We set the proportion of general to domain-specific data at 2:1.

Following DriveGPT4~\cite{xu2023drivegpt4}, we report several metric scores widely used in the NLP community, including CIDEr, BLEU4, and ROUGE-L, to evaluate the action descriptions and justifications. For open-loop control signal prediction, we use root mean squared error (RMSE) and threshold accuracies ($A_\tau$) for evaluation.

\begin{table}[t!]
  \centering
  \small
    \caption{\textbf{The results on action tasks of BDD-X dataset.} 
    We provide evaluation results on action description, action justification, and full-text generation (\ie, combining description and justification). ``B4'' stands for BLEU4. ``DA'' means model after domain adaptation on BDD-X.}
    \vspace{0.5em}
    \label{tab:results_on_bdd_action}
  \setlength{\tabcolsep}{3.7pt}
  \begin{tabular}{c|ccc|ccc|ccc}
   \toprule
    \multirow{2}{*}{Method}     & \multicolumn{3}{c|}{Description} & \multicolumn{3}{c|}{Justification} & \multicolumn{3}{c}{Full} \\
    & CIDEr & B4 &  ROUGE & CIDEr & B4 &  ROUGE & CIDEr & B4 &  ROUGE \\
    \midrule
     ADAPT~\cite{jin2023adapt} &219.35 &33.42&61.83&94.62&9.95&32.01&93.66&17.76&44.32 \\
     DriveGPT4~\cite{xu2023drivegpt4} &254.62 &35.99& 63.97 &101.55 &10.84 &31.91 &102.71 &19.00 &45.10\\

     \midrule
     \internvlbdd-1B &223.85&34.17&62.11&95.52&9.70&32.58&83.72&16.78&44.29 \\
     \internvlbdd-2B &242.14&35.77&63.03&105.06&10.63&32.46&98.47& 18.05 &44.52\\
     \internvlbdd-4B &237.41&35.94&63.67&104.62&9.51&32.23&97.42&17.70&44.98\\
   \bottomrule
  \end{tabular}

\end{table}
\begin{table}[t!]
  \small
  \centering
  \caption{\textbf{Quantitative results of control signals prediction on BDD-X test dataset.} RMSE denotes the root mean squared error, and $A_{\tau}$  measures the proportion of test samples with prediction errors less than $\tau$. ``DA'' means model after domain adaptation on BDD-X.}
  \vspace{0.5em}
  \label{tab:results_on_bdd_control}
  \setlength{\tabcolsep}{3.5pt}
  \begin{tabular}{c|ccccc|ccccc}
    \toprule        
    \multirow{2}{*}{Method}  &\multicolumn{5}{c|}{Speed(m/s)}&\multicolumn{5}{c}{Turningangle(degree)}\\
    &  RMSE$\downarrow$ &$A_{0.1}$$\uparrow$ & $A_{0.5}$$\uparrow$ & $A_{1.0}$$\uparrow$ & $A_{5.0}$$\uparrow$ &RMSE$\downarrow$ &$A_{0.1}$$\uparrow$ & $A_{0.5}$$\uparrow$& $A_{1.0}$$\uparrow$ & $A_{5.0}$$\uparrow$\\
    \midrule
    ADAPT~\cite{jin2023adapt} &3.02&9.56&24.77&37.07&90.39&11.98&27.93&66.83&75.13&89.45 \\ 
    DriveGPT4~\cite{xu2023drivegpt4} &1.30 &30.09 &60.88 &79.92& 98.44& 8.98 &59.23 &72.89 &79.59 &95.32 \\

     \midrule
     \internvlbdd-1B &1.28&29.44&60.38&79.34&98.67&9.45&59.34&73.54&80.28&92.76 \\
     \internvlbdd-2B &1.26&27.96&59.23&80.06&98.78 &9.52&57.40&72.54&80.06&92.04 \\
     \internvlbdd-4B &1.31 &28.84&60.94&78.78&98.61&9.46&59.12&73.15&80.17&92.65\\
  \bottomrule
  \end{tabular}

\end{table}
\paragraph{Results.}

We report our scores on the BDD-X testing set in \tabref{tab:results_on_bdd_action} and \tabref{tab:results_on_bdd_control}. Although our model has not undergone pre-training on large amounts of proprietary domain data like DriveGPT, it still performs comparably to DriveGPT4 across the four tasks and surpasses ADAPT~\cite{jin2023adapt}. Note that the performance of the three models on this dataset is similar, which aligns with the observations discussed in \secref{sec:Autonomous driving}.

\subsubsection{Medical Image Question Answering}
\label{sec:medical}

\begin{table}[t!]
  \centering
  \small
  \caption{\textbf{The sources of our medical data.} The table presents the data types and the sample sizes collected from each source. We sample a total of 500K image-text pairs from multiple publicly available datasets of different data types as our medical training data.}  
  \vspace{0.5em}
   \setlength{\tabcolsep}{8pt}
  \begin{tabular}{ccc}
     \toprule         
    Data       & Size  & Description\\
    \midrule
    PMC-VOA~\cite{lin2023pmc} & 238K & \\
    MedICaT~\cite{subramanian-2020-medicat} & 31K & The datasets include image-text pairs containing  \\
    PMC-Image~\cite{lin2023pmc,zhang2023pmc,wu2023towards} & 29K & X-rays, pathology images, and images of affected areas,\\
    Open-i~\cite{openi} & 1K & extracted from open-source websites or journal articles.\\
    MedPix~\cite{medpix} & 6K & \\
    \midrule
    \multirow{2}{*}{Quilt-1M~\cite{ikezogwo2023quilt}} & \multirow{2}{*}{95K} & A medical dataset includes image-text pairs of  \\
    && histopathology images.\\
    \midrule
    RP3D~\cite{wu2023towards} & 82K & \multirow{2}{*}{A medical dataset includes image-text pairs of X-ray images.} \\
    MIMIC-CXR~\cite{johnson2018mimic} & 14K & \\
    \midrule
    Retina Image Bank~\cite{imagebank} & 4K & A medical dataset includes image-text pairs of retinal images. \\
     \bottomrule
  \end{tabular}

  \label{tab:medical data}  

\end{table}

\begin{table}
  \centering
  \small
  \caption{\textbf{The results of our model on GMAI-MMBench.} The results of other models are taken from the GMAI-MMBench leaderboard. ``DA'' means model after domain adaptation on medical data. After supervised fine-tuning, our model shows significant improvement and outperforms several medical-specialized models (\eg, LLaVA-Med, RadFM) and some commercial closed-source models (\eg, Claude3-Opus) on most metrics. }
  \vspace{0.5em}
  \setlength{\tabcolsep}{3.5pt}
  \begin{tabular}{cccccccc}
    \toprule
    Model     & Size  & Seg C & Seg M & 2D Cls & 2D Det & 2D Mcls\_acc & 2D Mcls\_recall  \\
    \midrule
    Qwen-VL-Chat~\cite{Qwen-VL}	&9.6B &34.45  &	35.20 &	39.55	 &22.04	 &42.88	 &81.23 \\
    LLaVA-NeXT-mistral-7B~\cite{liu2023improved}	&7.6B	&	36.29	&	35.20&		39.34&		27.87&		44.05&		47.70 \\

    \midrule
    LLaVA-Med~\cite{li2023llavamed} & -	& 18.45	  & 18.97	 & 21.15	&17.14	&45.84	&41.19  \\
    RadFM~\cite{wu2023towards}	 &14B & 20.43	& 20.27& 	25.71& 	18.83	& 40.98& 	57.45 \\
    \midrule
    Claude3-Opus~\cite{claude3series2024}& -   & 33.56 & 33.36 & 32.17  &24.72   &45.31        &38.98   \\
    GPT-4V~\cite{gpt4v}	& -	& 	47.87& 		46.58& 		42.24& 		30.32	& 	45.21& 		40.59 \\
    \midrule
    \miniinternvl-1B &1B & 34.30&34.55& 36.02&24.08&21.67&8.57\\
    \miniinternvl-2B & 2B    & 35.33 & 35.61 & 38.08  &25.31   &43.52        &16.13   \\
    \miniinternvl-4B & 4B  & 36.60 & 36.99 & 38.74 & 26.01&43.99&16.25\\ 
    \internvlmi-1B  &1B& 38.67&39.44&35.87 &23.09&22.79&8.99 \\
    \internvlmi-2B &2B&40.22& 39.46 & 39.34  &25.59   &44.33 &16.20   \\
    \internvlmi-4B &4B &41.41&40.45 & 41.34 &24.84 &44.33&16.59 \\
    \bottomrule
  \end{tabular}

  \label{tab:medical results}  
\end{table}

\paragraph{Settings.}
We utilize several publicly available medical image-text datasets to improve the model's understanding of medical images. These datasets include a wide range of medical images, such as photos, X-rays, and pathology images. The datasets include PMC-OA~\cite{lin2023pmc}, MedICaT~\cite{subramanian-2020-medicat}, PMC-Image~\cite{lin2023pmc,zhang2023pmc,wu2023towards}, Open-i~\cite{openi}, MedPix~\cite{medpix}, Quilt-1M~\cite{ikezogwo2023quilt}, RP3D~\cite{wu2023towards}, MIMIC-CXR~\cite{johnson2018mimic}, and Retina Image Bank~\cite{imagebank}, which together provide a substantial collection of medical images. From these datasets, we sampled 500K image-text pairs for the model's training set. Finally,  We add general data in a 1:1 ratio to the domain-specific data and conduct full-parameter training of the model for one epoch.

\paragraph{Results.}
In this section, we present the performance of \miniinternvl and its fine-tuned variant, \textbf{\internvlmi }, on a comprehensive medical AI benchmark, GMAI-MMBench~\cite{chen2024gmai}.  Our evaluation is conducted using the VLMEvalKit\footnote{\url{https://github.com/open-compass/VLMEvalKit}} framework. \tabref{tab:medical results} illustrates the performance of various models on the medical VQA tasks.

After supervised fine-tuning, our model shows significant improvement across most evaluation metrics. Specifically, it excels in 2D classification (2D Cls), 2D detection (2D Det), and 2D multi-class accuracy (2D Mcls\_acc). These results highlight its strong multimodal understanding capabilities in complex medical visual question-answering tasks. 

Furthermore, our model of 4B size outperforms several medical-specialized models (\eg, LLaVA-Med~\cite{li2023llavamed}, RadFM~\cite{wu2023towards}) and some commercial closed-source models (\eg, Claude3-Opus~\cite{claude3series2024}) on most metrics. However, there is no improvement in multiple-choice questions after SFT, which we attribute to the lack of multiple-choice question data in the training dataset.

\subsubsection{Remote Sensing}
\label{sec: Remote}

\paragraph{Settings.} 
The training data is summarized in \tabref{tab: RS_Data}. The GeoChat instruction set~\cite{kuckreja2023geochat} serves as the primary component of our training dataset. To enrich the dataset with high-resolution imagery, we also incorporate the RSVQA-HR dataset~\cite{lobry2020rsvqa}. Additionally, we include 100K VQA instances sampled from FIT-RS~\cite{luo2024sky} to further expand our training set. For the visual grounding task, we integrate the DIOR-RSVG dataset~\cite{zhan2023rsvg} into our training process. All data are reformatted according to the methods outlined in \secref{sec: framework}.

A single epoch of training on the visual grounding data is found to be insufficient, so we repeat the DIOR-RSVG training samples multiple times. Finally, we incorporate 20\% of the total general domain training samples into the training data and conduct training following the settings described in \secref{sec:Autonomous driving}.

We assess the performance of our transferred model using the RSVQA dataset for the VQA task and the DIOR-RSVG dataset for the visual grounding task. Following the methodology outlined in~\cite{kuckreja2023geochat}, we chose the Presence, Comparison, and Rural/Urban subsets of the RSVQA-LR and RSVQA-HR datasets for assessment.

\begin{wraptable}{r}{0.55\textwidth}
  \centering
  \small
  \caption{\textbf{Details on the training samples used to adapt remote sensing.}}
  \setlength{\tabcolsep}{2pt}
  \begin{tabular}{c|cc}
    \toprule        
    Dataset & Data type  & Size  \\
    \midrule
    \multirow{6}{*}{GeoChat~\cite{kuckreja2023geochat}} & Detailed Description & 75K    \\
    &Multi-Round Conversation & 65K    \\
    &Complex Questions & 10K   \\
    &VQA & 91.5K \\
    &Region Captioning & 40K    \\
    &Visual Grounding& 25K  \\
    \midrule
    RSVQA-HR~\cite{lobry2020rsvqa} & VQA & 50K \\
    \midrule
    FIT-RS-VQA~\cite{luo2024sky} & VQA & 100K \\
    \midrule
    DIOR-RSVG~\cite{zhan2023rsvg} & Visual Grounding & 26K \\
    \bottomrule
  \end{tabular}
  \label{tab: RS_Data}  
\end{wraptable}

\begin{table}[t]
  \centering
  \caption{\textbf{The results on remote sensing VQA and visual grounding tasks.} For the VQA datasets, we omit area and count questions during evaluation.  ``DA'' means model after domain adaptation on remote sensing.  }
  \vspace{0.5em}
  \label{tab:results_on_rs} 
  \scriptsize
   \setlength{\tabcolsep}{1.5pt}
   \begin{tabular}{c|c|cccc|ccc|ccc|c}
    \toprule         
    \multirow{2}{*}{Method} & \multirow{2}{*}{Size}  &\multicolumn{4}{c|}{RSVQA-LR}&\multicolumn{3}{c|}{RSVQA-HR-Test1} & \multicolumn{3}{c|}{RSVQA-HR-Test2}& \ DIOR-RSVG \\
    & & Rural/Urban& Presence&Compare &Avg.&Presence &Compare& Avg. &Presence &Compare& Avg. & (acc@0.5) \\
    \midrule
    RSVQA~\cite{zhang2023spatial}         & -  &90.00 &87.46 &81.50 &86.32 &90.43 &88.19 &83.12 &86.26 &85.94 &77.50 & -  \\

    Bi-Modal~\cite{bazi2022bi}	   & -  &92.66 &91.06 &92.66 &91.63	&92.03 &91.83 &84.98 &89.37	&89.62 &80.54 & -  \\

    EasyToHard~\cite{yuan2022easy}	   & -  &91.67 &90.66 &87.49 &89.94	&91.39 &89.75 &93.97 &87.97	&87.68 &79.06 & -  \\[1pt]
    \midrule 
    GeoChat~\cite{kuckreja2023geochat}       & 7B &94.00 &91.09 &90.33  &90.70 &-  &-   &-   &- &58.45 & 83.19  &72.30 \\
    SkyEyeGPT~\cite{zhan2024skyeyegpt}     &-&75.00& 88.93&88.63& 84.19 & 84.95& 85.63&85.29&83.50& 80.28& 81.89&88.59 \\
    SkySenseGPT~\cite{luo2024sky} &-&95.00&91.07& 92.00& 92.69&-&-&-&69.14&84.14&76.64 & - \\[1pt]
    \midrule
    \miniinternvl-4B &4B& 66.00 &64.64&73.26&69.55 &62.42&79.20&71.72&65.73&79.70&73.55&16.89\\
    InternVL2-Llama3-76B &76B &61.00 &	66.29	&79.61	&73.77	&60.79	&77.47	&70.04	&63.30	&78.32	&71.71	&29.65\\
    \internvlrs-1B & 1B &95.00&81.39&91.6&87.37&92.24&91.76 &91.98&89.38&90.91&90.24&89.73\\ 
    \internvlrs-2B & 2B & 93.00 &87.07 &91.85 &89.87 &92.33&92.21&92.27&89.60&90.86&90.30&89.24 \\
    \internvlrs-4B & 4B &92.00 &85.69&92.18&89.46&92.42&92.12&92.25&89.25&90.92&90.18&92.04\\ 
    \bottomrule
  \end{tabular}
 
\end{table}

\paragraph{Results.} \tabref{tab:results_on_rs} presents the performance of our models on remote sensing VQA and visual grounding tasks. On the RSVQA task, our model demonstrates strong performance under both high-resolution and low-resolution conditions. Unlike existing remote sensing MLLMs such as GeoChat~\cite{kuckreja2023geochat} and SkySenseGPT~\cite{zhan2024skyeyegpt}, which support only single-resolution images, our model leverages dynamic resolution to effectively benefit from high-resolution training data. Compared to traditional models in the remote sensing domain--such as RSVQA~\cite{zhang2023spatial}, Bi-Modal~\cite{bazi2022bi}, and EasyToHard~\cite{yuan2022easy}--our model achieves superior scores on both RSVQA-HR-Test1 and RSVQA-HR-Test2, showcasing its generalization ability. Furthermore, our models of three different sizes outperform SkyEyeGPT on DIOR-RSVG, indicating that our framework can effectively model visual grounding tasks.

\subsection{Ablation Study}
\label{sec:ablation}

\begin{table}[t]
  \vspace{-0.5em}
  \centering
  \small
  \caption{\textbf{Comparison between \miniinternvl-2B with \miniinternvl-CLIP-2B across various tasks.} We test the model on DriveLM val set after fine-tuning them using domain-specific data.}
  \vspace{0.5em}
  \setlength{\tabcolsep}{2pt}
  \begin{tabular}{cccccccccccc}
    \toprule         
   \multirow{2}{*}{Method}  & MMB-& MMB- & ChartQA   & DocVQA  & InfoVQA   & \multirow{2}{*}{MMMU} & MME-RW&\multirow{2}{*}{DriveLM} \\
   & EN & CN & (test) &  (val) & (val) &&(AD)& \\ 
    \midrule
    \miniinternvl-CLIP-2B & 70.3&68.1  & 70.9  &  77.5  & 49.6  &  32.9  & 43.7  &0.580\\
    \miniinternvl-2B &  73.2&70.9  & 76.2  &  85.9 & 57.7  &  34.3 &  48.0 &0.578\\
    \bottomrule
  \end{tabular}

    \label{tab:ablation_clip}  
\end{table}
\begin{figure}[htbp]
    \centering
        \begin{subfigure}[b]{0.475\textwidth}
     \includegraphics[width=\textwidth]{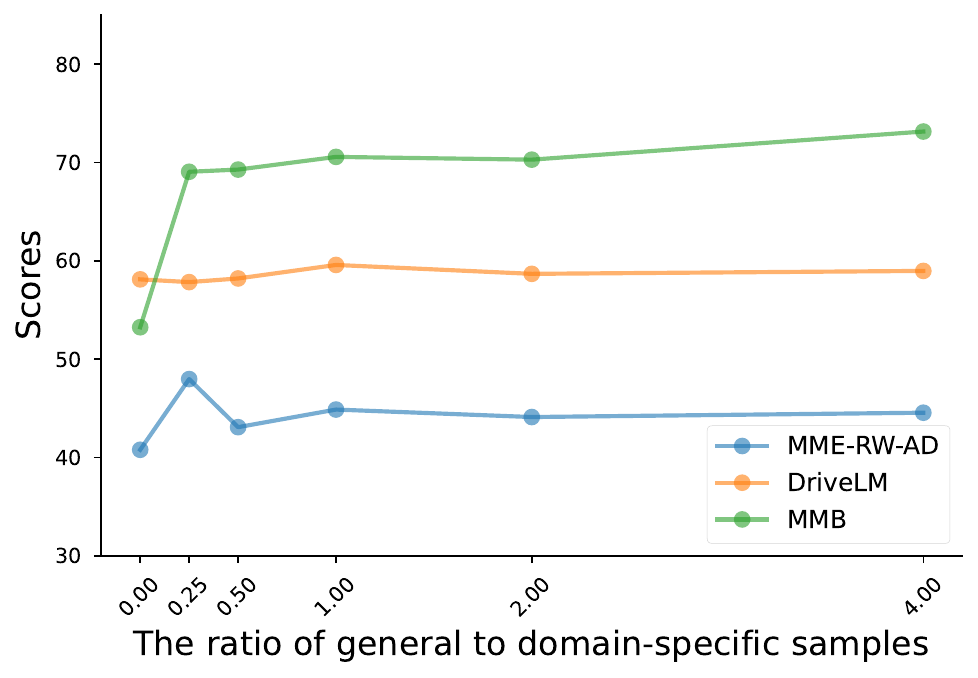}
     \caption{The impact of the proportion of general data.}
     \label{fig: data_ratio}
     \end{subfigure}
    \hfill
    \begin{subfigure}[b]{0.475\textwidth}
        \centering
           \includegraphics[width=\textwidth]{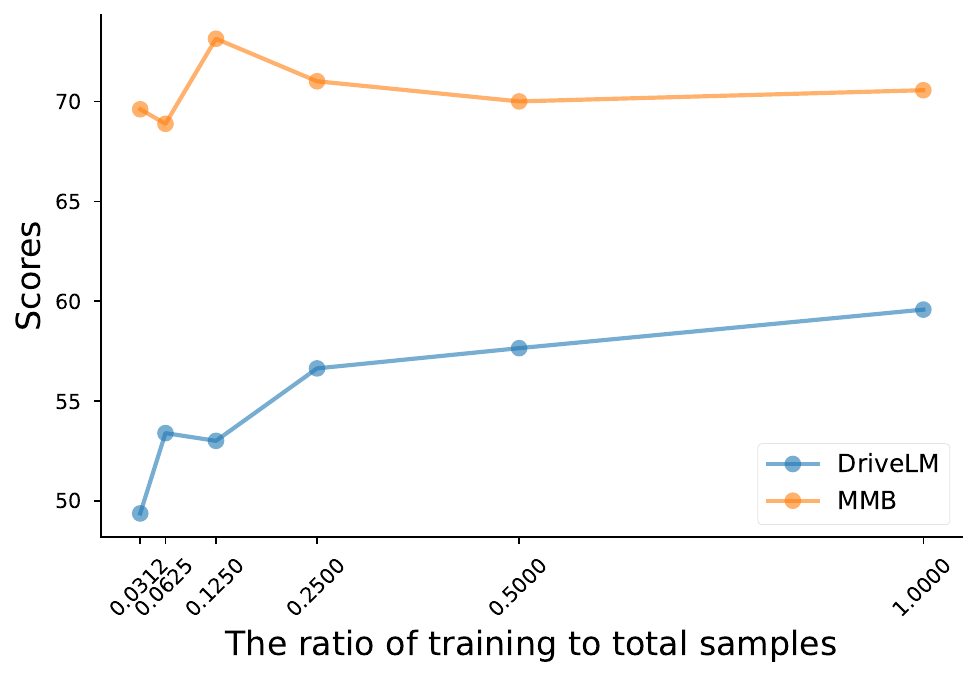}
        \caption{The impact of the number of training samples.}
        \label{fig: sample_size}
    \end{subfigure}

    \caption{\textbf{The impact of the proportion of general data and the number of training samples.}
    MME-RW-AD, MMB, and DriveLM refer to the Autonomous Driving domain of MME-RealWorld, the general multimodal benchmark MMBench, and DriveLM Challenge, respectively.
    }
    \label{fig: ablation}
\end{figure}

\begin{figure*}[ht]
  \centering
    \includegraphics[width=\textwidth]{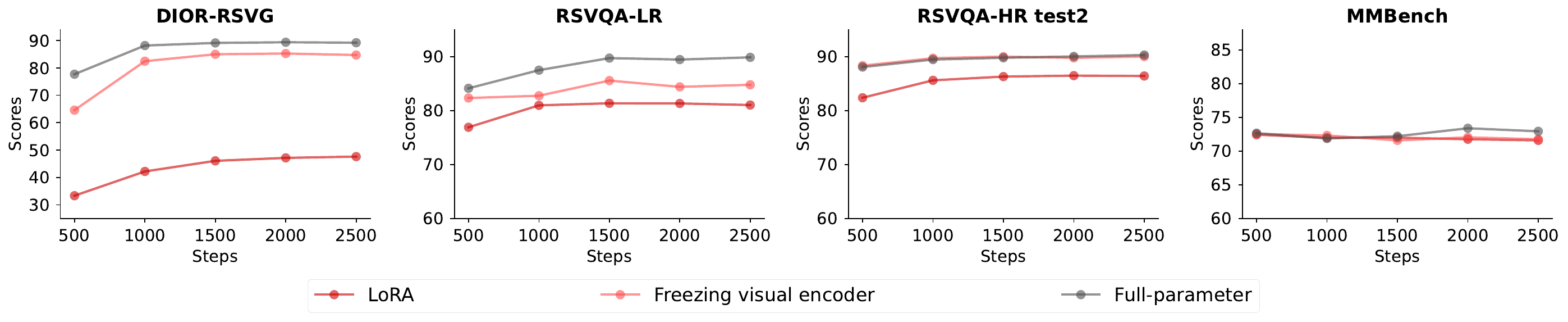}
  \caption{\textbf{Performance of 3 methods at varying training steps on 4 benchmarks.} After 1500 steps, the model's performance converges. The scores for the three adaptation methods remain stable across different training steps on the general multimodal benchmark MMBench~\cite{liu2023mmbench}.}
  \label{fig: ablation_steps}
\end{figure*}

\begin{wraptable}{r}{0.55\textwidth}
  \centering
  \small
  \vspace{-1.3em}
  \caption{\textbf{Comparing the per-GPU memory usage and training speed of 3 methods.} Our experiment use ZeRO1~\cite{rajbhandari2020zero} and is conducted on 8 A100 GPUs.}
  \setlength{\tabcolsep}{3pt}
  \begin{tabular}{c|cc}
    \toprule      
    \multirow{2}{*}{Method}  &GPU Memory & Training Speed \\ 
    &(GB)& (iterations per hour)\\
    \midrule
    LoRA & 23.40 & 263.02 \\
    Freezing ViT & 29.87 & 260.76\\
    Full-parameter & 33.11 & 185.13 \\
    \bottomrule
  \end{tabular}

    \label{tab:ablation_step}  
\end{wraptable}
\paragraph{The Importance of Knowledge Distillation.
}
We demonstrate the effectiveness of knowledge distillation in this study.
We train a new 2B-sized MLLM, which we refer to as \miniinternvl-CLIP-2B.
Specifically, we maintain the structure of the MLLM while replacing the vision encoder from InternViT with {CLIP-ViT-L-336px}~\cite{radford2021learning}, employing the same training recipe as described in \secref{sec: architecture}. 
As shown in \tabref{tab:ablation_clip}, the results across multiple benchmarks indicate that \miniinternvl-2B significantly outperforms \miniinternvl-CLIP-2B on general benchmarks, particularly in document-related tasks. This clearly illustrates that our knowledge distillation in our method effectively enables InternViT to acquire visual world knowledge. Furthermore, we adapt both models to autonomous driving tasks, and the results demonstrate the advantages of our model in proprietary domain transfer.

\paragraph{The Impact of Data Ratio Balancing.}

In this study, we investigate the effect of the proportion between general data and domain-specific data on model transferability. We conduct experiments in the autonomous driving domain, utilizing all samples from DriveLM and supplementing them with general training data at multiples of $r$ times the number of DriveLM samples. The results are shown in \figref{fig: data_ratio}. Our findings indicate that relying exclusively on domain-specific training data does not yield optimal performance on downstream tasks. Introducing a specific ratio of general data not only enhances performance on domain-specific tasks but also reduces performance degradation on general multimodal benchmarks. This demonstrates that incorporating an appropriate proportion of general data is crucial for improving the model's generalization ability and maintaining its general capabilities. For our autonomous driving scenario, we observe that performance peaks at  $ r = 0.25 $; beyond this point, performance slightly declines as  $ r  $ increases. This suggests that we can achieve benefits without significantly increasing computational load.

\paragraph{Influence of Training Sample Size.}

We investigate how varying the quantity of training samples affects performance on downstream tasks. In this experiment, we use different amounts of training data while maintaining a 1:1 ratio between general data and domain-specific data, as shown in \figref{fig: sample_size}. Training the model with only one-quarter of the full dataset significantly reduces the computational load during training while resulting in only a minor loss in performance. Notably, the model's score on general benchmarks remains largely unchanged when the proportion of different training data is kept constant.

\paragraph{Effect of Different Adaptation Methods.} 
In this study, we examine the effects of different adaptation methods--LoRA, freezing the vision encoder, and full-parameter fine-tuning--on model performance. Using the dataset described in \secref{sec: Remote}, we apply these three methods to train the model across varying numbers of steps and evaluate its performance on three tasks: general multimodal VQA, remote sensing VQA, and visual grounding. We record memory consumption and training speed during the process, as shown in \tabref{tab:ablation_step}. As illustrated in \figref{fig: ablation_steps}, model performance converges after 1500 steps, with each method exhibiting distinct performance ceilings. 
Notably, full-parameter fine-tuning achieves the highest scores on domain-specific tasks. Additionally, we find that LoRA underperforms on the visual grounding task, and freezing the vision encoder strikes a balance between performance and computational efficiency. 
The scores for all three adaptation methods remain stable across different training steps on the general multimodal benchmark, maintaining strong performance in the general domain even with extended training.

\section{Conclusion}
In this work, we introduce \miniinternvl, a series of lightweight, open-source MLLMs designed to tackle the challenges of deploying MLLMs in resource-constrained environments. 
\miniinternvl utilizes InternViT-300M as a compact vision encoder, integrating world knowledge across multiple domains through knowledge distillation from a more capable teacher model, thereby addressing the limitations of encoders like CLIP-ViT.
\miniinternvl achieves approximately 90\% of the performance of larger models using significantly fewer parameters, excelling particularly in tasks such as OCR and domain-specific image understanding.
To facilitate the application of small-scale multimodal models in specialized fields, we employ a unified transfer format, enabling our models to be effectively adapted to multiple specific domains, where they achieve comparable performance to other domain-specific approaches. We hope that this work provides valuable insights into the application of MLLM.

\newpage
\bibliographystyle{unsrtnat}
\bibliography{citation}

\begin{thebibliography}{151}
\providecommand{\natexlab}[1]{#1}
\providecommand{\url}[1]{\texttt{#1}}
\expandafter\ifx\csname urlstyle\endcsname\relax
  \providecommand{\doi}[1]{doi: #1}\else
  \providecommand{\doi}{doi: \begingroup \urlstyle{rm}\Url}\fi

\bibitem[Chen et~al.(2024{\natexlab{a}})Chen, Wu, Wang, Su, Chen, Xing, Zhong, Zhang, Zhu, Lu, et~al.]{chen2023internvl}
Zhe Chen, Jiannan Wu, Wenhai Wang, Weijie Su, Guo Chen, Sen Xing, Muyan Zhong, Qinglong Zhang, Xizhou Zhu, Lewei Lu, et~al.
\newblock Internvl: Scaling up vision foundation models and aligning for generic visual-linguistic tasks.
\newblock In \emph{CVPR}, pages 24185--24198, 2024{\natexlab{a}}.

\bibitem[Chen et~al.(2024{\natexlab{b}})Chen, Wang, Tian, Ye, Gao, Cui, Tong, Hu, Luo, Ma, et~al.]{chen2024far}
Zhe Chen, Weiyun Wang, Hao Tian, Shenglong Ye, Zhangwei Gao, Erfei Cui, Wenwen Tong, Kongzhi Hu, Jiapeng Luo, Zheng Ma, et~al.
\newblock How far are we to gpt-4v? closing the gap to commercial multimodal models with open-source suites.
\newblock \emph{arXiv preprint arXiv:2404.16821}, 2024{\natexlab{b}}.

\bibitem[OpenAI(2023)]{gpt4v}
OpenAI.
\newblock Gpt-4v(ision) system card.
\newblock \url{https://cdn.openai.com/papers/GPTV_System_Card.pdf}, 2023.

\bibitem[OpenAI(2024)]{openai2024gpt4o}
OpenAI.
\newblock Hello gpt-4o.
\newblock \url{https://openai.com/index/hello-gpt-4o/}, 2024.

\bibitem[Li et~al.(2024{\natexlab{a}})Li, Zhang, Guo, Zhang, Li, Zhang, Zhang, Li, Liu, and Li]{li2024llava}
Bo~Li, Yuanhan Zhang, Dong Guo, Renrui Zhang, Feng Li, Hao Zhang, Kaichen Zhang, Yanwei Li, Ziwei Liu, and Chunyuan Li.
\newblock Llava-onevision: Easy visual task transfer.
\newblock \emph{arXiv preprint arXiv:2408.03326}, 2024{\natexlab{a}}.

\bibitem[Liu et~al.(2024{\natexlab{a}})Liu, Li, Li, Li, Zhang, Shen, and Lee]{liu2024llavanext}
Haotian Liu, Chunyuan Li, Yuheng Li, Bo~Li, Yuanhan Zhang, Sheng Shen, and Yong~Jae Lee.
\newblock Llava-next: Improved reasoning, ocr, and world knowledge, January 2024{\natexlab{a}}.
\newblock URL \url{https://llava-vl.github.io/blog/2024-01-30-llava-next/}.

\bibitem[Wang et~al.(2024{\natexlab{a}})Wang, Zhang, Luo, Sun, Cui, Wang, Zhang, Wang, Li, Yu, et~al.]{wang2024emu3}
Xinlong Wang, Xiaosong Zhang, Zhengxiong Luo, Quan Sun, Yufeng Cui, Jinsheng Wang, Fan Zhang, Yueze Wang, Zhen Li, Qiying Yu, et~al.
\newblock Emu3: Next-token prediction is all you need.
\newblock \emph{arXiv preprint arXiv:2409.18869}, 2024{\natexlab{a}}.

\bibitem[Lin et~al.(2024{\natexlab{a}})Lin, Yin, Ping, Molchanov, Shoeybi, and Han]{lin2024vila}
Ji~Lin, Hongxu Yin, Wei Ping, Pavlo Molchanov, Mohammad Shoeybi, and Song Han.
\newblock Vila: On pre-training for visual language models.
\newblock In \emph{CVPR}, pages 26689--26699, 2024{\natexlab{a}}.

\bibitem[Ye et~al.(2024)Ye, Xu, Liu, Hu, Yan, Qian, Zhang, Huang, and Zhou]{ye2024mplug}
Jiabo Ye, Haiyang Xu, Haowei Liu, Anwen Hu, Ming Yan, Qi~Qian, Ji~Zhang, Fei Huang, and Jingren Zhou.
\newblock mplug-owl3: Towards long image-sequence understanding in multi-modal large language models.
\newblock \emph{arXiv preprint arXiv:2408.04840}, 2024.

\bibitem[Ge et~al.(2024)Ge, Cheng, Wang, Yuan, Gao, Song, Song, Huang, and Zheng]{ge2024convllava}
Chunjiang Ge, Sijie Cheng, Ziming Wang, Jiale Yuan, Yuan Gao, Jun Song, Shiji Song, Gao Huang, and Bo~Zheng.
\newblock Convllava: Hierarchical backbones as visual encoder for large multimodal models.
\newblock \emph{arXiv preprint arXiv:2405.15738}, 2024.

\bibitem[Shi et~al.(2024)Shi, Liu, Wang, Liao, Radhakrishnan, Huang, Yin, Sapra, Yacoob, Shi, et~al.]{shi2024eagle}
Min Shi, Fuxiao Liu, Shihao Wang, Shijia Liao, Subhashree Radhakrishnan, De-An Huang, Hongxu Yin, Karan Sapra, Yaser Yacoob, Humphrey Shi, et~al.
\newblock Eagle: Exploring the design space for multimodal llms with mixture of encoders.
\newblock \emph{arXiv preprint arXiv:2408.15998}, 2024.

\bibitem[Dubey et~al.(2024)Dubey, Jauhri, Pandey, Kadian, Al-Dahle, Letman, Mathur, Schelten, Yang, Fan, et~al.]{llama3modelcard}
Abhimanyu Dubey, Abhinav Jauhri, Abhinav Pandey, Abhishek Kadian, Ahmad Al-Dahle, Aiesha Letman, Akhil Mathur, Alan Schelten, Amy Yang, Angela Fan, et~al.
\newblock The llama 3 herd of models.
\newblock \emph{arXiv preprint arXiv:2407.21783}, 2024.

\bibitem[Touvron et~al.(2023)Touvron, Martin, Stone, Albert, Almahairi, Babaei, Bashlykov, Batra, Bhargava, Bhosale, et~al.]{touvron2023llama2}
Hugo Touvron, Louis Martin, Kevin Stone, Peter Albert, Amjad Almahairi, Yasmine Babaei, Nikolay Bashlykov, Soumya Batra, Prajjwal Bhargava, Shruti Bhosale, et~al.
\newblock Llama 2: Open foundation and fine-tuned chat models.
\newblock \emph{arXiv preprint arXiv:2307.09288}, 2023.

\bibitem[Cai et~al.(2024)Cai, Cao, Chen, Chen, Chen, Chen, Chen, Chen, Chen, Chu, et~al.]{cai2024internlm2}
Zheng Cai, Maosong Cao, Haojiong Chen, Kai Chen, Keyu Chen, Xin Chen, Xun Chen, Zehui Chen, Zhi Chen, Pei Chu, et~al.
\newblock Internlm2 technical report.
\newblock \emph{arXiv preprint arXiv:2403.17297}, 2024.

\bibitem[Bai et~al.(2023{\natexlab{a}})Bai, Bai, Chu, Cui, Dang, Deng, Fan, Ge, Han, Huang, et~al.]{qwen}
Jinze Bai, Shuai Bai, Yunfei Chu, Zeyu Cui, Kai Dang, Xiaodong Deng, Yang Fan, Wenbin Ge, Yu~Han, Fei Huang, et~al.
\newblock Qwen technical report.
\newblock \emph{arXiv preprint arXiv:2309.16609}, 2023{\natexlab{a}}.

\bibitem[Yang et~al.(2024)Yang, Yang, Hui, Zheng, Yu, Zhou, Li, Li, Liu, Huang, et~al.]{qwen2}
An~Yang, Baosong Yang, Binyuan Hui, Bo~Zheng, Bowen Yu, Chang Zhou, Chengpeng Li, Chengyuan Li, Dayiheng Liu, Fei Huang, et~al.
\newblock Qwen2 technical report.
\newblock \emph{arXiv preprint arXiv:2407.10671}, 2024.

\bibitem[Jiang et~al.(2023)Jiang, Sablayrolles, Mensch, Bamford, Chaplot, Casas, Bressand, Lengyel, Lample, Saulnier, et~al.]{jiang2023mistral}
Albert~Q Jiang, Alexandre Sablayrolles, Arthur Mensch, Chris Bamford, Devendra~Singh Chaplot, Diego de~las Casas, Florian Bressand, Gianna Lengyel, Guillaume Lample, Lucile Saulnier, et~al.
\newblock Mistral 7b.
\newblock \emph{arXiv preprint arXiv:2310.06825}, 2023.

\bibitem[Radford et~al.(2021)Radford, Kim, Hallacy, Ramesh, Goh, Agarwal, Sastry, Askell, Mishkin, and Clark]{radford2021learning}
Alec Radford, Jong~Wook Kim, Chris Hallacy, Aditya Ramesh, Gabriel Goh, Sandhini Agarwal, Girish Sastry, Amanda Askell, Pamela Mishkin, and Jack Clark.
\newblock Learning transferable visual models from natural language supervision.
\newblock In \emph{ICML}, 2021.

\bibitem[Zhai et~al.(2023)Zhai, Mustafa, Kolesnikov, and Beyer]{zhai2023sigmoid}
Xiaohua Zhai, Basil Mustafa, Alexander Kolesnikov, and Lucas Beyer.
\newblock Sigmoid loss for language image pre-training.
\newblock In \emph{ICCV}, pages 11975--11986, 2023.

\bibitem[Yao et~al.(2024)Yao, Yu, Zhang, Wang, Cui, Zhu, Cai, Li, Zhao, He, et~al.]{yao2024minicpm}
Yuan Yao, Tianyu Yu, Ao~Zhang, Chongyi Wang, Junbo Cui, Hongji Zhu, Tianchi Cai, Haoyu Li, Weilin Zhao, Zhihui He, et~al.
\newblock Minicpm-v: A gpt-4v level mllm on your phone.
\newblock \emph{arXiv preprint arXiv:2408.01800}, 2024.

\bibitem[Li et~al.(2023{\natexlab{a}})Li, Zhang, Wang, Zhong, Chen, Chu, Liu, and Jia]{li2024mgm}
Yanwei Li, Yuechen Zhang, Chengyao Wang, Zhisheng Zhong, Yixin Chen, Ruihang Chu, Shaoteng Liu, and Jiaya Jia.
\newblock Mini-gemini: Mining the potential of multi-modality vision language models.
\newblock \emph{arXiv:2403.18814}, 2023{\natexlab{a}}.

\bibitem[Chu et~al.(2023)Chu, Qiao, Lin, Xu, Yang, Hu, Wei, Zhang, Zhang, Wei, et~al.]{chu2023mobilevlm}
Xiangxiang Chu, Limeng Qiao, Xinyang Lin, Shuang Xu, Yang Yang, Yiming Hu, Fei Wei, Xinyu Zhang, Bo~Zhang, Xiaolin Wei, et~al.
\newblock Mobilevlm: A fast, reproducible and strong vision language assistant for mobile devices.
\newblock \emph{arXiv preprint arXiv:2312.16886}, 2023.

\bibitem[Chu et~al.(2024)Chu, Qiao, Zhang, Xu, Wei, Yang, Sun, Hu, Lin, Zhang, et~al.]{chu2024mobilevlm}
Xiangxiang Chu, Limeng Qiao, Xinyu Zhang, Shuang Xu, Fei Wei, Yang Yang, Xiaofei Sun, Yiming Hu, Xinyang Lin, Bo~Zhang, et~al.
\newblock Mobilevlm v2: Faster and stronger baseline for vision language model.
\newblock \emph{arXiv preprint arXiv:2402.03766}, 2024.

\bibitem[Beyer et~al.(2024)Beyer, Steiner, Pinto, Kolesnikov, Wang, Salz, Neumann, Alabdulmohsin, Tschannen, Bugliarello, et~al.]{beyer2024paligemma}
Lucas Beyer, Andreas Steiner, Andr{\'e}~Susano Pinto, Alexander Kolesnikov, Xiao Wang, Daniel Salz, Maxim Neumann, Ibrahim Alabdulmohsin, Michael Tschannen, Emanuele Bugliarello, et~al.
\newblock Paligemma: A versatile 3b vlm for transfer.
\newblock \emph{arXiv preprint arXiv:2407.07726}, 2024.

\bibitem[He et~al.(2024)He, Liu, Wu, Yuan, Wang, Huang, and Zhao]{he2024bunny}
Muyang He, Yexin Liu, Boya Wu, Jianhao Yuan, Yueze Wang, Tiejun Huang, and Bo~Zhao.
\newblock Efficient multimodal learning from data-centric perspective.
\newblock \emph{arXiv preprint arXiv:2402.11530}, 2024.

\bibitem[Devlin(2018)]{devlin2018bert}
Jacob Devlin.
\newblock Bert: Pre-training of deep bidirectional transformers for language understanding.
\newblock \emph{arXiv preprint arXiv:1810.04805}, 2018.

\bibitem[Jia et~al.(2021)Jia, Yang, Xia, Chen, Parekh, Pham, Le, Sung, Li, and Duerig]{jia2021scaling}
Chao Jia, Yinfei Yang, Ye~Xia, Yi-Ting Chen, Zarana Parekh, Hieu Pham, Quoc Le, Yun-Hsuan Sung, Zhen Li, and Tom Duerig.
\newblock Scaling up visual and vision-language representation learning with noisy text supervision.
\newblock In \emph{ICML}, pages 4904--4916. PMLR, 2021.

\bibitem[Abdin et~al.(2024)Abdin, Jacobs, Awan, Aneja, Awadallah, Awadalla, Bach, Bahree, Bakhtiari, Behl, et~al.]{abdin2024phi}
Marah Abdin, Sam~Ade Jacobs, Ammar~Ahmad Awan, Jyoti Aneja, Ahmed Awadallah, Hany Awadalla, Nguyen Bach, Amit Bahree, Arash Bakhtiari, Harkirat Behl, et~al.
\newblock Phi-3 technical report: A highly capable language model locally on your phone.
\newblock \emph{arXiv preprint arXiv:2404.14219}, 2024.

\bibitem[Liu et~al.(2023{\natexlab{a}})Liu, Duan, Zhang, Li, Zhang, Zhao, Yuan, Wang, He, Liu, et~al.]{liu2023mmbench}
Yuan Liu, Haodong Duan, Yuanhan Zhang, Bo~Li, Songyang Zhang, Wangbo Zhao, Yike Yuan, Jiaqi Wang, Conghui He, Ziwei Liu, et~al.
\newblock Mmbench: Is your multi-modal model an all-around player?
\newblock \emph{arXiv preprint arXiv:2307.06281}, 2023{\natexlab{a}}.

\bibitem[Masry et~al.(2022{\natexlab{a}})Masry, Do, Tan, Joty, and Hoque]{masry2022chartqa}
Ahmed Masry, Xuan~Long Do, Jia~Qing Tan, Shafiq Joty, and Enamul Hoque.
\newblock Chartqa: A benchmark for question answering about charts with visual and logical reasoning.
\newblock In \emph{ACL}, pages 2263--2279, 2022{\natexlab{a}}.

\bibitem[Lu et~al.(2023{\natexlab{a}})Lu, Bansal, Xia, Liu, Li, Hajishirzi, Cheng, Chang, Galley, and Gao]{lu2023mathvista}
Pan Lu, Hritik Bansal, Tony Xia, Jiacheng Liu, Chunyuan Li, Hannaneh Hajishirzi, Hao Cheng, Kai-Wei Chang, Michel Galley, and Jianfeng Gao.
\newblock Mathvista: Evaluating mathematical reasoning of foundation models in visual contexts.
\newblock \emph{arXiv preprint arXiv:2310.02255}, 2023{\natexlab{a}}.

\bibitem[Liu et~al.(2023{\natexlab{b}})Liu, He, Wang, Wang, Wang, Chen, Zhang, Lai, Yang, Li, et~al.]{liu2023interngpt}
Zhaoyang Liu, Yinan He, Wenhai Wang, Weiyun Wang, Yi~Wang, Shoufa Chen, Qinglong Zhang, Zeqiang Lai, Yang Yang, Qingyun Li, et~al.
\newblock Interngpt: Solving vision-centric tasks by interacting with chatgpt beyond language.
\newblock \emph{arXiv preprint arXiv:2305.05662}, 2023{\natexlab{b}}.

\bibitem[Shen et~al.(2024)Shen, Song, Tan, Li, Lu, and Zhuang]{shen2024hugginggpt}
Yongliang Shen, Kaitao Song, Xu~Tan, Dongsheng Li, Weiming Lu, and Yueting Zhuang.
\newblock Hugginggpt: Solving ai tasks with chatgpt and its friends in hugging face.
\newblock \emph{NeurIPS}, 36, 2024.

\bibitem[Wu et~al.(2023{\natexlab{a}})Wu, Yin, Qi, Wang, Tang, and Duan]{wu2023visualgpt}
Chenfei Wu, Shengming Yin, Weizhen Qi, Xiaodong Wang, Zecheng Tang, and Nan Duan.
\newblock Visual chatgpt: Talking, drawing and editing with visual foundation models.
\newblock \emph{arXiv preprint arXiv:2303.04671}, 2023{\natexlab{a}}.

\bibitem[Li et~al.(2022)Li, Li, Xiong, and Hoi]{li2022blip}
Junnan Li, Dongxu Li, Caiming Xiong, and Steven Hoi.
\newblock Blip: Bootstrapping language-image pre-training for unified vision-language understanding and generation.
\newblock In \emph{ICML}, pages 12888--12900. PMLR, 2022.

\bibitem[Li et~al.(2023{\natexlab{b}})Li, Li, Savarese, and Hoi]{li2023blip2}
Junnan Li, Dongxu Li, Silvio Savarese, and Steven Hoi.
\newblock Blip-2: Bootstrapping language-image pre-training with frozen image encoders and large language models.
\newblock In \emph{ICML}, pages 19730--19742. PMLR, 2023{\natexlab{b}}.

\bibitem[Liu et~al.(2024{\natexlab{b}})Liu, Li, Wu, and Lee]{liu2023llava}
Haotian Liu, Chunyuan Li, Qingyang Wu, and Yong~Jae Lee.
\newblock Visual instruction tuning.
\newblock \emph{NeurIPS}, 36, 2024{\natexlab{b}}.

\bibitem[Bai et~al.(2023{\natexlab{b}})Bai, Bai, Yang, Wang, Tan, Wang, Lin, Zhou, and Zhou]{Qwen-VL}
Jinze Bai, Shuai Bai, Shusheng Yang, Shijie Wang, Sinan Tan, Peng Wang, Junyang Lin, Chang Zhou, and Jingren Zhou.
\newblock Qwen-vl: A versatile vision-language model for understanding, localization, text reading, and beyond.
\newblock \emph{arXiv preprint arXiv:2308.12966}, 2023{\natexlab{b}}.

\bibitem[Peng et~al.(2023)Peng, Wang, Dong, Hao, Huang, Ma, and Wei]{peng2023kosmos2}
Zhiliang Peng, Wenhui Wang, Li~Dong, Yaru Hao, Shaohan Huang, Shuming Ma, and Furu Wei.
\newblock Kosmos-2: Grounding multimodal large language models to the world.
\newblock \emph{arXiv preprint arXiv:2306.14824}, 2023.

\bibitem[Wang et~al.(2024{\natexlab{b}})Wang, Zhang, Ren, Duan, Li, Liu, Hu, Chen, Zhang, Lu, et~al.]{wang2024mmniah}
Weiyun Wang, Shuibo Zhang, Yiming Ren, Yuchen Duan, Tiantong Li, Shuo Liu, Mengkang Hu, Zhe Chen, Kaipeng Zhang, Lewei Lu, et~al.
\newblock Needle in a multimodal haystack.
\newblock \emph{arXiv preprint arXiv:2406.07230}, 2024{\natexlab{b}}.

\bibitem[Alayrac et~al.(2022)Alayrac, Donahue, Luc, Miech, Barr, Hasson, Lenc, Mensch, Millican, Reynolds, et~al.]{alayrac2022flamingo}
Jean-Baptiste Alayrac, Jeff Donahue, Pauline Luc, Antoine Miech, Iain Barr, Yana Hasson, Karel Lenc, Arthur Mensch, Katherine Millican, Malcolm Reynolds, et~al.
\newblock Flamingo: a visual language model for few-shot learning.
\newblock \emph{NeurIPS}, 35:\penalty0 23716--23736, 2022.

\bibitem[Wang et~al.(2023{\natexlab{a}})Wang, Lv, Yu, Hong, Qi, Wang, Ji, Yang, Zhao, Song, et~al.]{wang2023cogvlm}
Weihan Wang, Qingsong Lv, Wenmeng Yu, Wenyi Hong, Ji~Qi, Yan Wang, Junhui Ji, Zhuoyi Yang, Lei Zhao, Xixuan Song, et~al.
\newblock Cogvlm: Visual expert for pretrained language models.
\newblock \emph{arXiv preprint arXiv:2311.03079}, 2023{\natexlab{a}}.

\bibitem[Tian et~al.(2024{\natexlab{a}})Tian, Zhu, Xiong, Wang, Chen, Wang, Chen, Lu, Lu, Zhou, et~al.]{tian2024mminterleaved}
Changyao Tian, Xizhou Zhu, Yuwen Xiong, Weiyun Wang, Zhe Chen, Wenhai Wang, Yuntao Chen, Lewei Lu, Tong Lu, Jie Zhou, et~al.
\newblock Mm-interleaved: Interleaved image-text generative modeling via multi-modal feature synchronizer.
\newblock \emph{arXiv preprint arXiv:2401.10208}, 2024{\natexlab{a}}.

\bibitem[Bavishi et~al.(2023)Bavishi, Elsen, Hawthorne, Nye, Odena, Somani, and Ta\c{s}\i{}rlar]{fuyu-8b}
Rohan Bavishi, Erich Elsen, Curtis Hawthorne, Maxwell Nye, Augustus Odena, Arushi Somani, and Sa\u{g}nak Ta\c{s}\i{}rlar.
\newblock Introducing our multimodal models, 2023.
\newblock URL \url{https://www.adept.ai/blog/fuyu-8b}.

\bibitem[Lin et~al.(2024{\natexlab{b}})Lin, Shrivastava, Luo, Iyer, Lewis, Gosh, Zettlemoyer, and Aghajanyan]{lin2024moma}
Xi~Victoria Lin, Akshat Shrivastava, Liang Luo, Srinivasan Iyer, Mike Lewis, Gargi Gosh, Luke Zettlemoyer, and Armen Aghajanyan.
\newblock Moma: Efficient early-fusion pre-training with mixture of modality-aware experts.
\newblock \emph{arXiv preprint arXiv:2407.21770}, 2024{\natexlab{b}}.

\bibitem[Team(2024)]{team2024chameleon}
Chameleon Team.
\newblock Chameleon: Mixed-modal early-fusion foundation models.
\newblock \emph{arXiv preprint arXiv:2405.09818}, 2024.

\bibitem[Sun et~al.(2023)Sun, Fang, Wu, Wang, and Cao]{EVA-CLIP}
Quan Sun, Yuxin Fang, Ledell Wu, Xinlong Wang, and Yue Cao.
\newblock Eva-clip: Improved training techniques for clip at scale.
\newblock \emph{arXiv preprint arXiv:2303.15389}, 2023.

\bibitem[Tong et~al.(2024{\natexlab{a}})Tong, Liu, Zhai, Ma, LeCun, and Xie]{tong2024eyes}
Shengbang Tong, Zhuang Liu, Yuexiang Zhai, Yi~Ma, Yann LeCun, and Saining Xie.
\newblock Eyes wide shut? exploring the visual shortcomings of multimodal llms.
\newblock In \emph{CVPR}, pages 9568--9578, 2024{\natexlab{a}}.

\bibitem[Oquab et~al.(2023)Oquab, Darcet, Moutakanni, Vo, Szafraniec, Khalidov, Fernandez, Haziza, Massa, El-Nouby, et~al.]{oquab2023dinov2}
Maxime Oquab, Timoth{\'e}e Darcet, Th{\'e}o Moutakanni, Huy Vo, Marc Szafraniec, Vasil Khalidov, Pierre Fernandez, Daniel Haziza, Francisco Massa, Alaaeldin El-Nouby, et~al.
\newblock Dinov2: Learning robust visual features without supervision.
\newblock \emph{arXiv preprint arXiv:2304.07193}, 2023.

\bibitem[Luo et~al.(2024{\natexlab{a}})Luo, Zhou, Zhang, Zheng, Sun, and Ji]{luo2024feast}
Gen Luo, Yiyi Zhou, Yuxin Zhang, Xiawu Zheng, Xiaoshuai Sun, and Rongrong Ji.
\newblock Feast your eyes: Mixture-of-resolution adaptation for multimodal large language models.
\newblock \emph{arXiv preprint arXiv:2403.03003}, 2024{\natexlab{a}}.

\bibitem[Lu et~al.(2024)Lu, Liu, Zhang, Wang, Dong, Liu, Sun, Ren, Li, Sun, et~al.]{lu2024deepseekvl}
Haoyu Lu, Wen Liu, Bo~Zhang, Bingxuan Wang, Kai Dong, Bo~Liu, Jingxiang Sun, Tongzheng Ren, Zhuoshu Li, Yaofeng Sun, et~al.
\newblock Deepseek-vl: towards real-world vision-language understanding.
\newblock \emph{arXiv preprint arXiv:2403.05525}, 2024.

\bibitem[Kirillov et~al.(2023{\natexlab{a}})Kirillov, Mintun, Ravi, Mao, Rolland, Gustafson, Xiao, Whitehead, Berg, Lo, Doll{\'a}r, and Girshick]{kirillov2023segany}
Alexander Kirillov, Eric Mintun, Nikhila Ravi, Hanzi Mao, Chloe Rolland, Laura Gustafson, Tete Xiao, Spencer Whitehead, Alexander~C. Berg, Wan-Yen Lo, Piotr Doll{\'a}r, and Ross Girshick.
\newblock Segment anything.
\newblock \emph{arXiv:2304.02643}, 2023{\natexlab{a}}.

\bibitem[Kuckreja et~al.(2024)Kuckreja, Danish, Naseer, Das, Khan, and Khan]{kuckreja2023geochat}
Kartik Kuckreja, Muhammad~S. Danish, Muzammal Naseer, Abhijit Das, Salman Khan, and Fahad~S. Khan.
\newblock Geochat: Grounded large vision-language model for remote sensing.
\newblock \emph{CVPR}, 2024.

\bibitem[Zhang et~al.(2024{\natexlab{a}})Zhang, Cai, Zhang, Zhuang, and Mao]{zhang2024earthgpt}
Wei Zhang, Miaoxin Cai, Tong Zhang, Yin Zhuang, and Xuerui Mao.
\newblock Earthgpt: A universal multi-modal large language model for multi-sensor image comprehension in remote sensing domain.
\newblock \emph{TGRS}, 2024{\natexlab{a}}.

\bibitem[Li et~al.(2023{\natexlab{c}})Li, Wong, Zhang, Usuyama, Liu, Yang, Naumann, Poon, and Gao]{li2023llavamed}
Chunyuan Li, Cliff Wong, Sheng Zhang, Naoto Usuyama, Haotian Liu, Jianwei Yang, Tristan Naumann, Hoifung Poon, and Jianfeng Gao.
\newblock Llava-med: Training a large language-and-vision assistant for biomedicine in one day.
\newblock \emph{arXiv preprint arXiv:2306.00890}, 2023{\natexlab{c}}.

\bibitem[Liu et~al.(2023{\natexlab{c}})Liu, Wang, Ye, Chong, Zhou, and Hua]{Liu2023QilinMedVLTC}
Junling Liu, Ziming Wang, Qichen Ye, Dading Chong, Peilin Zhou, and Yining Hua.
\newblock Qilin-med-vl: Towards chinese large vision-language model for general healthcare.
\newblock \emph{arXiv preprint arXiv:2310.17956}, 2023{\natexlab{c}}.

\bibitem[Li et~al.(2024{\natexlab{b}})Li, Zhang, Wang, Hao, Lei, Tan, Zhou, Liu, Wang, Chen, et~al.]{li2024chemvlm}
Junxian Li, Di~Zhang, Xunzhi Wang, Zeying Hao, Jingdi Lei, Qian Tan, Cai Zhou, Wei Liu, Weiyun Wang, Zhe Chen, et~al.
\newblock Seeing and understanding: Bridging vision with chemical knowledge via chemvlm.
\newblock \emph{arXiv preprint arXiv:2408.07246}, 2024{\natexlab{b}}.

\bibitem[Tian et~al.(2024{\natexlab{b}})Tian, Gu, Li, Liu, Hu, Wang, Zhan, Jia, Lang, and Zhao]{tian2024drivevlm}
Xiaoyu Tian, Junru Gu, Bailin Li, Yicheng Liu, Chenxu Hu, Yang Wang, Kun Zhan, Peng Jia, Xianpeng Lang, and Hang Zhao.
\newblock Drivevlm: The convergence of autonomous driving and large vision-language models.
\newblock \emph{arXiv preprint arXiv:2402.12289}, 2024{\natexlab{b}}.

\bibitem[Wang et~al.(2023{\natexlab{b}})Wang, Xie, Hu, Zou, Fan, Tong, Wen, Wu, Deng, Li, et~al.]{wang2023drivemlm}
Wenhai Wang, Jiangwei Xie, ChuanYang Hu, Haoming Zou, Jianan Fan, Wenwen Tong, Yang Wen, Silei Wu, Hanming Deng, Zhiqi Li, et~al.
\newblock Drivemlm: Aligning multi-modal large language models with behavioral planning states for autonomous driving.
\newblock \emph{arXiv preprint arXiv:2312.09245}, 2023{\natexlab{b}}.

\bibitem[Xu et~al.(2023)Xu, Zhang, Xie, Zhao, Guo, Wong, Li, and Zhao]{xu2023drivegpt4}
Zhenhua Xu, Yujia Zhang, Enze Xie, Zhen Zhao, Yong Guo, Kwan-Yee~K. Wong, Zhenguo Li, and Hengshuang Zhao.
\newblock Drivegpt4: Interpretable end-to-end autonomous driving via large language model.
\newblock \emph{arXiv preprint arXiv:2310.01412}, 2023.

\bibitem[Wang et~al.(2024{\natexlab{c}})Wang, Shi, Li, Wang, Huang, Xing, Chen, Li, Zhu, Cao, et~al.]{wang2023allseeing}
Weiyun Wang, Min Shi, Qingyun Li, Wenhai Wang, Zhenhang Huang, Linjie Xing, Zhe Chen, Hao Li, Xizhou Zhu, Zhiguo Cao, et~al.
\newblock The all-seeing project: Towards panoptic visual recognition and understanding of the open world.
\newblock In \emph{ICLR}, 2024{\natexlab{c}}.

\bibitem[Wang et~al.(2024{\natexlab{d}})Wang, Ren, Luo, Li, Yan, Chen, Wang, Li, Lu, Zhu, et~al.]{wang2024allseeingv2}
Weiyun Wang, Yiming Ren, Haowen Luo, Tiantong Li, Chenxiang Yan, Zhe Chen, Wenhai Wang, Qingyun Li, Lewei Lu, Xizhou Zhu, et~al.
\newblock The all-seeing project v2: Towards general relation comprehension of the open world.
\newblock \emph{arXiv preprint arXiv:2402.19474}, 2024{\natexlab{d}}.

\bibitem[Schuhmann et~al.(2022{\natexlab{a}})Schuhmann, Beaumont, Vencu, Gordon, Wightman, Cherti, Coombes, Katta, Mullis, Wortsman, et~al.]{schuhmann2022laion5b}
Christoph Schuhmann, Romain Beaumont, Richard Vencu, Cade Gordon, Ross Wightman, Mehdi Cherti, Theo Coombes, Aarush Katta, Clayton Mullis, Mitchell Wortsman, et~al.
\newblock Laion-5b: An open large-scale dataset for training next generation image-text models.
\newblock \emph{NeurIPS}, 35:\penalty0 25278--25294, 2022{\natexlab{a}}.

\bibitem[Byeon et~al.(2022)Byeon, Park, Kim, Lee, Baek, and Kim]{byeon2022coyo}
Minwoo Byeon, Beomhee Park, Haecheon Kim, Sungjun Lee, Woonhyuk Baek, and Saehoon Kim.
\newblock Coyo-700m: Image-text pair dataset, 2022.

\bibitem[Chen et~al.(2015)Chen, Fang, Lin, Vedantam, Gupta, Doll{\'a}r, and Zitnick]{chen2015cococaption}
Xinlei Chen, Hao Fang, Tsung-Yi Lin, Ramakrishna Vedantam, Saurabh Gupta, Piotr Doll{\'a}r, and C~Lawrence Zitnick.
\newblock Microsoft coco captions: Data collection and evaluation server.
\newblock \emph{arXiv preprint arXiv:1504.00325}, 2015.

\bibitem[Gupta et~al.(2019)Gupta, Dollar, and Girshick]{gupta2019lvis}
Agrim Gupta, Piotr Dollar, and Ross Girshick.
\newblock {LVIS}: A dataset for large vocabulary instance segmentation.
\newblock In \emph{CVPR}, 2019.

\bibitem[Shao et~al.(2019)Shao, Li, Zhang, Peng, Yu, Zhang, Li, and Sun]{shao2019objects365}
Shuai Shao, Zeming Li, Tianyuan Zhang, Chao Peng, Gang Yu, Xiangyu Zhang, Jing Li, and Jian Sun.
\newblock Objects365: A large-scale, high-quality dataset for object detection.
\newblock In \emph{ICCV}, pages 8430--8439, 2019.

\bibitem[Plummer et~al.(2015)Plummer, Wang, Cervantes, Caicedo, Hockenmaier, and Lazebnik]{plummer2015flickr30k}
Bryan~A Plummer, Liwei Wang, Chris~M Cervantes, Juan~C Caicedo, Julia Hockenmaier, and Svetlana Lazebnik.
\newblock Flickr30k entities: Collecting region-to-phrase correspondences for richer image-to-sentence models.
\newblock In \emph{ICCV}, pages 2641--2649, 2015.

\bibitem[Krishna et~al.(2017)Krishna, Zhu, Groth, Johnson, Hata, Kravitz, Chen, Kalantidis, Li, Shamma, et~al.]{krishna2017visual}
Ranjay Krishna, Yuke Zhu, Oliver Groth, Justin Johnson, Kenji Hata, Joshua Kravitz, Stephanie Chen, Yannis Kalantidis, Li-Jia Li, David~A Shamma, et~al.
\newblock Visual genome: Connecting language and vision using crowdsourced dense image annotations.
\newblock \emph{IJCV}, 123:\penalty0 32--73, 2017.

\bibitem[Liu et~al.(2024{\natexlab{c}})Liu, Cao, Gao, Wang, Chen, Wang, Tian, Lu, Zhu, Lu, et~al.]{liu2024mminstruct}
Yangzhou Liu, Yue Cao, Zhangwei Gao, Weiyun Wang, Zhe Chen, Wenhai Wang, Hao Tian, Lewei Lu, Xizhou Zhu, Tong Lu, et~al.
\newblock Mminstruct: A high-quality multi-modal instruction tuning dataset with extensive diversity.
\newblock \emph{arXiv preprint arXiv:2407.15838}, 2024{\natexlab{c}}.

\bibitem[Liu et~al.(2023{\natexlab{d}})Liu, Lin, Li, Wang, Yacoob, and Wang]{liu2023aligning}
Fuxiao Liu, Kevin Lin, Linjie Li, Jianfeng Wang, Yaser Yacoob, and Lijuan Wang.
\newblock Aligning large multi-modal model with robust instruction tuning.
\newblock \emph{arXiv preprint arXiv:2306.14565}, 2023{\natexlab{d}}.

\bibitem[Sidorov et~al.(2020)Sidorov, Hu, Rohrbach, and Singh]{sidorov2020textcaps}
Oleksii Sidorov, Ronghang Hu, Marcus Rohrbach, and Amanpreet Singh.
\newblock Textcaps: a dataset for image captioning with reading comprehension.
\newblock In \emph{ECCV}, pages 742--758, 2020.

\bibitem[Gu et~al.(2022)Gu, Meng, Lu, Hou, Minzhe, Liang, Yao, Huang, Zhang, Jiang, et~al.]{gu2022wukong}
Jiaxi Gu, Xiaojun Meng, Guansong Lu, Lu~Hou, Niu Minzhe, Xiaodan Liang, Lewei Yao, Runhui Huang, Wei Zhang, Xin Jiang, et~al.
\newblock Wukong: A 100 million large-scale chinese cross-modal pre-training benchmark.
\newblock \emph{NeurIPS}, 35:\penalty0 26418--26431, 2022.

\bibitem[Yuan et~al.(2019)Yuan, Zhu, Xu, Li, Mu, and Hu]{yuan2019ctw}
Tai-Ling Yuan, Zhe Zhu, Kun Xu, Cheng-Jun Li, Tai-Jiang Mu, and Shi-Min Hu.
\newblock A large chinese text dataset in the wild.
\newblock \emph{Journal of Computer Science and Technology}, 34:\penalty0 509--521, 2019.

\bibitem[Liu et~al.(2023{\natexlab{e}})Liu, Wang, Yao, Chen, Song, Cho, Yacoob, and Yu]{liu2023mmcinst}
Fuxiao Liu, Xiaoyang Wang, Wenlin Yao, Jianshu Chen, Kaiqiang Song, Sangwoo Cho, Yaser Yacoob, and Dong Yu.
\newblock Mmc: Advancing multimodal chart understanding with large-scale instruction tuning.
\newblock \emph{arXiv preprint arXiv:2311.10774}, 2023{\natexlab{e}}.

\bibitem[Sun et~al.(2019)Sun, Ni, Chng, Liu, Luo, Ng, Han, Ding, Liu, Karatzas, et~al.]{sun2019lsvt}
Yipeng Sun, Zihan Ni, Chee-Kheng Chng, Yuliang Liu, Canjie Luo, Chun~Chet Ng, Junyu Han, Errui Ding, Jingtuo Liu, Dimosthenis Karatzas, et~al.
\newblock Icdar 2019 competition on large-scale street view text with partial labeling-rrc-lsvt.
\newblock In \emph{ICDAR}, pages 1557--1562, 2019.

\bibitem[Biten et~al.(2019)Biten, Tito, Mafla, Gomez, Rusinol, Valveny, Jawahar, and Karatzas]{biten2019stvqa}
Ali~Furkan Biten, Ruben Tito, Andres Mafla, Lluis Gomez, Mar{\c{c}}al Rusinol, Ernest Valveny, CV~Jawahar, and Dimosthenis Karatzas.
\newblock Scene text visual question answering.
\newblock In \emph{ICCV}, pages 4291--4301, 2019.

\bibitem[Shi et~al.(2017)Shi, Yao, Liao, Yang, Xu, Cui, Belongie, Lu, and Bai]{shi2017rctw17}
Baoguang Shi, Cong Yao, Minghui Liao, Mingkun Yang, Pei Xu, Linyan Cui, Serge Belongie, Shijian Lu, and Xiang Bai.
\newblock Icdar2017 competition on reading chinese text in the wild (rctw-17).
\newblock In \emph{ICDAR}, volume~1, pages 1429--1434, 2017.

\bibitem[Zhang et~al.(2019)Zhang, Zhou, Jiang, Song, Li, Zhou, Wang, Wang, Liao, Yang, et~al.]{zhang2019rects}
Rui Zhang, Yongsheng Zhou, Qianyi Jiang, Qi~Song, Nan Li, Kai Zhou, Lei Wang, Dong Wang, Minghui Liao, Mingkun Yang, et~al.
\newblock Icdar 2019 robust reading challenge on reading chinese text on signboard.
\newblock In \emph{ICDAR}, pages 1577--1581, 2019.

\bibitem[Chng et~al.(2019)Chng, Liu, Sun, Ng, Luo, Ni, Fang, Zhang, Han, Ding, et~al.]{chng2019art}
Chee~Kheng Chng, Yuliang Liu, Yipeng Sun, Chun~Chet Ng, Canjie Luo, Zihan Ni, ChuanMing Fang, Shuaitao Zhang, Junyu Han, Errui Ding, et~al.
\newblock Icdar2019 robust reading challenge on arbitrary-shaped text-rrc-art.
\newblock In \emph{ICDAR}, pages 1571--1576, 2019.

\bibitem[Kim et~al.(2022{\natexlab{a}})Kim, Hong, Yim, Nam, Park, Yim, Hwang, Yun, Han, and Park]{kim2022synthdog}
Geewook Kim, Teakgyu Hong, Moonbin Yim, JeongYeon Nam, Jinyoung Park, Jinyeong Yim, Wonseok Hwang, Sangdoo Yun, Dongyoon Han, and Seunghyun Park.
\newblock Ocr-free document understanding transformer.
\newblock In \emph{ECCV}, 2022{\natexlab{a}}.

\bibitem[Schuhmann et~al.(2022{\natexlab{b}})Schuhmann, Köpf, Vencu, Coombes, and Beaumont]{schuhmann2022laioncoco}
Christoph Schuhmann, Andreas Köpf, Richard Vencu, Theo Coombes, and Romain Beaumont.
\newblock Laion coco: 600m synthetic captions from laion2b-en.
\newblock \emph{https://laion.ai/blog/laion-coco/}, 2022{\natexlab{b}}.

\bibitem[Veit et~al.(2016)Veit, Matera, Neumann, Matas, and Belongie]{veit2016cocotext}
Andreas Veit, Tomas Matera, Lukas Neumann, Jiri Matas, and Serge Belongie.
\newblock Coco-text: Dataset and benchmark for text detection and recognition in natural images.
\newblock \emph{arXiv preprint arXiv:1601.07140}, 2016.

\bibitem[Mathew et~al.(2021)Mathew, Karatzas, and Jawahar]{mathew2021docvqa}
Minesh Mathew, Dimosthenis Karatzas, and CV~Jawahar.
\newblock Docvqa: A dataset for vqa on document images.
\newblock In \emph{WACV}, pages 2200--2209, 2021.

\bibitem[Singh et~al.(2021)Singh, Pang, Toh, Huang, Galuba, and Hassner]{singh2021textocr}
Amanpreet Singh, Guan Pang, Mandy Toh, Jing Huang, Wojciech Galuba, and Tal Hassner.
\newblock Textocr: Towards large-scale end-to-end reasoning for arbitrary-shaped scene text.
\newblock In \emph{CVPR}, pages 8802--8812, 2021.

\bibitem[Zhang et~al.(2023{\natexlab{a}})Zhang, Zhang, Gu, Zhou, Lipka, Yang, and Sun]{zhang2023llavar}
Yanzhe Zhang, Ruiyi Zhang, Jiuxiang Gu, Yufan Zhou, Nedim Lipka, Diyi Yang, and Tong Sun.
\newblock Llavar: Enhanced visual instruction tuning for text-rich image understanding.
\newblock \emph{arXiv preprint arXiv:2306.17107}, 2023{\natexlab{a}}.

\bibitem[Kembhavi et~al.(2017)Kembhavi, Seo, Schwenk, Choi, Farhadi, and Hajishirzi]{kembhavi2017you}
Aniruddha Kembhavi, Minjoon Seo, Dustin Schwenk, Jonghyun Choi, Ali Farhadi, and Hannaneh Hajishirzi.
\newblock Are you smarter than a sixth grader? textbook question answering for multimodal machine comprehension.
\newblock In \emph{CVPR}, pages 4999--5007, 2017.

\bibitem[Gupta et~al.(2016)Gupta, Vedaldi, and Zisserman]{Gupta16}
Ankush Gupta, Andrea Vedaldi, and Andrew Zisserman.
\newblock Synthetic data for text localisation in natural images.
\newblock In \emph{CVPR}, pages 2315--2324, 2016.

\bibitem[Hu et~al.(2024)Hu, Xu, Ye, Yan, Zhang, Zhang, Li, Zhang, Jin, Huang, et~al.]{hu2024mplug}
Anwen Hu, Haiyang Xu, Jiabo Ye, Ming Yan, Liang Zhang, Bo~Zhang, Chen Li, Ji~Zhang, Qin Jin, Fei Huang, et~al.
\newblock mplug-docowl 1.5: Unified structure learning for ocr-free document understanding.
\newblock \emph{arXiv preprint arXiv:2403.12895}, 2024.

\bibitem[Kembhavi et~al.(2016{\natexlab{a}})Kembhavi, Salvato, Kolve, Seo, Hajishirzi, and Farhadi]{kembhavi2016ai2d}
Aniruddha Kembhavi, Mike Salvato, Eric Kolve, Minjoon Seo, Hannaneh Hajishirzi, and Ali Farhadi.
\newblock A diagram is worth a dozen images.
\newblock In \emph{ECCV}, pages 235--251, 2016{\natexlab{a}}.

\bibitem[Methani et~al.(2020)Methani, Ganguly, Khapra, and Kumar]{methani2020plotqa}
Nitesh Methani, Pritha Ganguly, Mitesh~M Khapra, and Pratyush Kumar.
\newblock Plotqa: Reasoning over scientific plots.
\newblock In \emph{WACV}, pages 1527--1536, 2020.

\bibitem[Mathew et~al.(2022)Mathew, Bagal, Tito, Karatzas, Valveny, and Jawahar]{mathew2022infographicvqa}
Minesh Mathew, Viraj Bagal, Rub{\`e}n Tito, Dimosthenis Karatzas, Ernest Valveny, and CV~Jawahar.
\newblock Infographicvqa.
\newblock In \emph{WACV}, pages 1697--1706, 2022.

\bibitem[Chang et~al.(2022)Chang, Palzer, Li, Fosler-Lussier, and Xiao]{chang2022mapqa}
Shuaichen Chang, David Palzer, Jialin Li, Eric Fosler-Lussier, and Ningchuan Xiao.
\newblock Mapqa: A dataset for question answering on choropleth maps.
\newblock In \emph{NeurIPS Workshop}, 2022.

\bibitem[Kahou et~al.(2017)Kahou, Michalski, Atkinson, K{\'a}d{\'a}r, Trischler, and Bengio]{kahou2017figureqa}
Samira~Ebrahimi Kahou, Vincent Michalski, Adam Atkinson, {\'A}kos K{\'a}d{\'a}r, Adam Trischler, and Yoshua Bengio.
\newblock Figureqa: An annotated figure dataset for visual reasoning.
\newblock \emph{arXiv preprint arXiv:1710.07300}, 2017.

\bibitem[Lu et~al.(2021{\natexlab{a}})Lu, Qiu, Chen, Xia, Zhao, Zhang, Yu, Liang, and Zhu]{lu2021iconqa}
Pan Lu, Liang Qiu, Jiaqi Chen, Tony Xia, Yizhou Zhao, Wei Zhang, Zhou Yu, Xiaodan Liang, and Song-Chun Zhu.
\newblock Iconqa: A new benchmark for abstract diagram understanding and visual language reasoning.
\newblock In \emph{NeurIPS}, 2021{\natexlab{a}}.

\bibitem[Liu et~al.(2023{\natexlab{f}})Liu, Wang, Yao, Chen, Song, Cho, Yacoob, and Yu]{liu2023mmc}
Fuxiao Liu, Xiaoyang Wang, Wenlin Yao, Jianshu Chen, Kaiqiang Song, Sangwoo Cho, Yaser Yacoob, and Dong Yu.
\newblock Mmc: Advancing multimodal chart understanding with large-scale instruction tuning.
\newblock \emph{arXiv preprint arXiv:2311.10774}, 2023{\natexlab{f}}.

\bibitem[Li et~al.(2023{\natexlab{d}})Li, Wang, Stengel-Eskin, Kortylewski, Ma, Van~Durme, and Yuille]{li2023superclevr}
Zhuowan Li, Xingrui Wang, Elias Stengel-Eskin, Adam Kortylewski, Wufei Ma, Benjamin Van~Durme, and Alan~L Yuille.
\newblock Super-clevr: A virtual benchmark to diagnose domain robustness in visual reasoning.
\newblock In \emph{CVPR}, pages 14963--14973, 2023{\natexlab{d}}.

\bibitem[Lindstr{\"o}m and Abraham(2022)]{lindstrom2022clevrmath}
Adam~Dahlgren Lindstr{\"o}m and Savitha~Sam Abraham.
\newblock Clevr-math: A dataset for compositional language, visual and mathematical reasoning.
\newblock \emph{arXiv preprint arXiv:2208.05358}, 2022.

\bibitem[Cao and Xiao(2022{\natexlab{a}})]{cao2022augmented}
Jie Cao and Jing Xiao.
\newblock An augmented benchmark dataset for geometric question answering through dual parallel text encoding.
\newblock In \emph{COLING}, pages 1511--1520, 2022{\natexlab{a}}.

\bibitem[Masry et~al.(2023)Masry, Kavehzadeh, Do, Hoque, and Joty]{masry2023unichart}
Ahmed Masry, Parsa Kavehzadeh, Xuan~Long Do, Enamul Hoque, and Shafiq Joty.
\newblock Unichart: A universal vision-language pretrained model for chart comprehension and reasoning.
\newblock \emph{arXiv preprint arXiv:2305.14761}, 2023.

\bibitem[Lu et~al.(2022)Lu, Mishra, Xia, Qiu, Chang, Zhu, Tafjord, Clark, and Kalyan]{lu2022scienceqa}
Pan Lu, Swaroop Mishra, Tanglin Xia, Liang Qiu, Kai-Wei Chang, Song-Chun Zhu, Oyvind Tafjord, Peter Clark, and Ashwin Kalyan.
\newblock Learn to explain: Multimodal reasoning via thought chains for science question answering.
\newblock \emph{NeurIPS}, 35:\penalty0 2507--2521, 2022.

\bibitem[Lu et~al.(2021{\natexlab{b}})Lu, Gong, Jiang, Qiu, Huang, Liang, and Zhu]{lu2021inter}
Pan Lu, Ran Gong, Shibiao Jiang, Liang Qiu, Siyuan Huang, Xiaodan Liang, and Song-Chun Zhu.
\newblock Inter-gps: Interpretable geometry problem solving with formal language and symbolic reasoning.
\newblock In \emph{ACL}, 2021{\natexlab{b}}.

\bibitem[Chen et~al.(2022)Chen, Li, Qin, Lu, Lin, Chen, and Liang]{chen2022unigeo}
Jiaqi Chen, Tong Li, Jinghui Qin, Pan Lu, Liang Lin, Chongyu Chen, and Xiaodan Liang.
\newblock Unigeo: Unifying geometry logical reasoning via reformulating mathematical expression.
\newblock \emph{arXiv preprint arXiv:2212.02746}, 2022.

\bibitem[Zhang et~al.(2023{\natexlab{b}})Zhang, Wu, Zhao, Lin, Zhang, Wang, and Xie]{zhang2023pmc}
Xiaoman Zhang, Chaoyi Wu, Ziheng Zhao, Weixiong Lin, Ya~Zhang, Yanfeng Wang, and Weidi Xie.
\newblock Pmc-vqa: Visual instruction tuning for medical visual question answering.
\newblock \emph{arXiv preprint arXiv:2305.10415}, 2023{\natexlab{b}}.

\bibitem[Lu et~al.(2023{\natexlab{b}})Lu, Qiu, Chang, Wu, Zhu, Rajpurohit, Clark, and Kalyan]{lu2023dynamic}
Pan Lu, Liang Qiu, Kai-Wei Chang, Ying~Nian Wu, Song-Chun Zhu, Tanmay Rajpurohit, Peter Clark, and Ashwin Kalyan.
\newblock Dynamic prompt learning via policy gradient for semi-structured mathematical reasoning.
\newblock In \emph{ICLR}, 2023{\natexlab{b}}.

\bibitem[Yu et~al.(2023)Yu, Jiang, Shi, Yu, Liu, Zhang, Kwok, Li, Weller, and Liu]{yu2023metamath}
Longhui Yu, Weisen Jiang, Han Shi, Jincheng Yu, Zhengying Liu, Yu~Zhang, James~T Kwok, Zhenguo Li, Adrian Weller, and Weiyang Liu.
\newblock Metamath: Bootstrap your own mathematical questions for large language models.
\newblock \emph{arXiv preprint arXiv:2309.12284}, 2023.

\bibitem[Yao et~al.(2011)Yao, Jiang, Khosla, Lin, Guibas, and Fei-Fei]{yao2011human}
Bangpeng Yao, Xiaoye Jiang, Aditya Khosla, Andy~Lai Lin, Leonidas Guibas, and Li~Fei-Fei.
\newblock Human action recognition by learning bases of action attributes and parts.
\newblock In \emph{ICCV}, pages 1331--1338, 2011.

\bibitem[Hudson and Manning(2019)]{hudson2019gqa}
Drew~A Hudson and Christopher~D Manning.
\newblock Gqa: A new dataset for real-world visual reasoning and compositional question answering.
\newblock In \emph{CVPR}, pages 6700--6709, 2019.

\bibitem[Huang et~al.(2020)Huang, Xiong, Rao, Wang, and Lin]{huang2020movienet}
Qingqiu Huang, Yu~Xiong, Anyi Rao, Jiaze Wang, and Dahua Lin.
\newblock Movienet: A holistic dataset for movie understanding.
\newblock In \emph{ECCV}, 2020.

\bibitem[{Hosu} et~al.(2020){Hosu}, {Lin}, {Sziranyi}, and {Saupe}]{koniq10k}
V.~{Hosu}, H.~{Lin}, T.~{Sziranyi}, and D.~{Saupe}.
\newblock Koniq-10k: An ecologically valid database for deep learning of blind image quality assessment.
\newblock \emph{TIP}, 29:\penalty0 4041--4056, 2020.

\bibitem[Mao et~al.(2017)Mao, Cheung, and She]{mao2017deepart}
Hui Mao, Ming Cheung, and James She.
\newblock Deepart: Learning joint representations of visual arts.
\newblock In \emph{ACM MM}, pages 1183--1191. ACM, 2017.

\bibitem[Lerner et~al.(2022)Lerner, Ferret, Guinaudeau, Le~Borgne, Besan{\c{c}}on, Moreno, and Lov{\'o}n~Melgarejo]{lerner2022viquae}
Paul Lerner, Olivier Ferret, Camille Guinaudeau, Herv{\'e} Le~Borgne, Romaric Besan{\c{c}}on, Jos{\'e}~G Moreno, and Jes{\'u}s Lov{\'o}n~Melgarejo.
\newblock Viquae, a dataset for knowledge-based visual question answering about named entities.
\newblock In \emph{SIGIR}, pages 3108--3120, 2022.

\bibitem[Lobry et~al.(2020)Lobry, Marcos, Murray, and Tuia]{lobry2020rsvqa}
Sylvain Lobry, Diego Marcos, Jesse Murray, and Devis Tuia.
\newblock Rsvqa: Visual question answering for remote sensing data.
\newblock \emph{TGRS}, 58\penalty0 (12):\penalty0 8555--8566, 2020.

\bibitem[Sima et~al.(2023)Sima, Renz, Chitta, Chen, Zhang, Xie, Luo, Geiger, and Li]{sima2023drivelm}
Chonghao Sima, Katrin Renz, Kashyap Chitta, Li~Chen, Hanxue Zhang, Chengen Xie, Ping Luo, Andreas Geiger, and Hongyang Li.
\newblock Drivelm: Driving with graph visual question answering.
\newblock \emph{arXiv preprint arXiv:2312.14150}, 2023.

\bibitem[Duan et~al.(2024)Duan, Yang, Qiao, Fang, Chen, Liu, Dong, Zang, Zhang, Wang, et~al.]{duan2024vlmevalkit}
Haodong Duan, Junming Yang, Yuxuan Qiao, Xinyu Fang, Lin Chen, Yuan Liu, Xiaoyi Dong, Yuhang Zang, Pan Zhang, Jiaqi Wang, et~al.
\newblock Vlmevalkit: An open-source toolkit for evaluating large multi-modality models.
\newblock \emph{arXiv preprint arXiv:2407.11691}, 2024.

\bibitem[Liu et~al.(2023{\natexlab{g}})Liu, Li, Li, Yu, Huang, Peng, Liu, Chen, Li, Jin, et~al.]{liu2023ocrbench}
Yuliang Liu, Zhang Li, Hongliang Li, Wenwen Yu, Mingxin Huang, Dezhi Peng, Mingyu Liu, Mingrui Chen, Chunyuan Li, Lianwen Jin, et~al.
\newblock On the hidden mystery of ocr in large multimodal models.
\newblock \emph{arXiv preprint arXiv:2305.07895}, 2023{\natexlab{g}}.

\bibitem[Yue et~al.(2024)Yue, Ni, Zhang, Zheng, Liu, Zhang, Stevens, Jiang, Ren, Sun, et~al.]{yue2023mmmu}
Xiang Yue, Yuansheng Ni, Kai Zhang, Tianyu Zheng, Ruoqi Liu, Ge~Zhang, Samuel Stevens, Dongfu Jiang, Weiming Ren, Yuxuan Sun, et~al.
\newblock Mmmu: A massive multi-discipline multimodal understanding and reasoning benchmark for expert agi.
\newblock In \emph{CVPR}, pages 9556--9567, 2024.

\bibitem[Achiam et~al.(2023)Achiam, Adler, Agarwal, Ahmad, Akkaya, Aleman, Almeida, Altenschmidt, Altman, Anadkat, et~al.]{openai2023gpt4}
Josh Achiam, Steven Adler, Sandhini Agarwal, Lama Ahmad, Ilge Akkaya, Florencia~Leoni Aleman, Diogo Almeida, Janko Altenschmidt, Sam Altman, Shyamal Anadkat, et~al.
\newblock Gpt-4 technical report.
\newblock \emph{arXiv preprint arXiv:2303.08774}, 2023.

\bibitem[Reid et~al.(2024)Reid, Savinov, Teplyashin, Lepikhin, Lillicrap, Alayrac, Soricut, Lazaridou, Firat, Schrittwieser, et~al.]{reid2024gemini1_5}
Machel Reid, Nikolay Savinov, Denis Teplyashin, Dmitry Lepikhin, Timothy Lillicrap, Jean-baptiste Alayrac, Radu Soricut, Angeliki Lazaridou, Orhan Firat, Julian Schrittwieser, et~al.
\newblock Gemini 1.5: Unlocking multimodal understanding across millions of tokens of context.
\newblock \emph{arXiv preprint arXiv:2403.05530}, 2024.

\bibitem[{Anthropic}(2024{\natexlab{a}})]{claude3_5series2024}
{Anthropic}.
\newblock Claude 3.5 sonnet model card addendum.
\newblock \url{https://www.anthropic.com}, 2024{\natexlab{a}}.
\newblock URL \url{https://www-cdn.anthropic.com/fed9cc193a14b84131812372d8d5857f8f304c52/Model_Card_Claude_3_Addendum.pdf}.

\bibitem[Tong et~al.(2024{\natexlab{b}})Tong, Brown, Wu, Woo, Middepogu, Akula, Yang, Yang, Iyer, Pan, et~al.]{tong2024cambrian1}
Shengbang Tong, Ellis Brown, Penghao Wu, Sanghyun Woo, Manoj Middepogu, Sai~Charitha Akula, Jihan Yang, Shusheng Yang, Adithya Iyer, Xichen Pan, et~al.
\newblock Cambrian-1: A fully open, vision-centric exploration of multimodal llms.
\newblock \emph{arXiv preprint arXiv:2406.16860}, 2024{\natexlab{b}}.

\bibitem[Wang et~al.(2024{\natexlab{e}})Wang, Bai, Tan, Wang, Fan, Bai, Chen, Liu, Wang, Ge, Fan, Dang, Du, Ren, Men, Liu, Zhou, Zhou, and Lin]{Qwen2VL}
Peng Wang, Shuai Bai, Sinan Tan, Shijie Wang, Zhihao Fan, Jinze Bai, Keqin Chen, Xuejing Liu, Jialin Wang, Wenbin Ge, Yang Fan, Kai Dang, Mengfei Du, Xuancheng Ren, Rui Men, Dayiheng Liu, Chang Zhou, Jingren Zhou, and Junyang Lin.
\newblock Qwen2-vl: Enhancing vision-language model's perception of the world at any resolution.
\newblock \emph{arXiv preprint arXiv:2409.12191}, 2024{\natexlab{e}}.

\bibitem[Chen et~al.(2023)Chen, Li, Dong, Zhang, He, Wang, Zhao, and Lin]{chen2023sharegpt4v}
Lin Chen, Jisong Li, Xiaoyi Dong, Pan Zhang, Conghui He, Jiaqi Wang, Feng Zhao, and Dahua Lin.
\newblock Sharegpt4v: Improving large multi-modal models with better captions.
\newblock \emph{arXiv preprint arXiv:2311.12793}, 2023.

\bibitem[Kafle et~al.(2018)Kafle, Price, Cohen, and Kanan]{kafle2018dvqa}
Kushal Kafle, Brian Price, Scott Cohen, and Christopher Kanan.
\newblock Dvqa: Understanding data visualizations via question answering.
\newblock In \emph{CVPR}, pages 5648--5656, 2018.

\bibitem[Masry et~al.(2022{\natexlab{b}})Masry, Do, Tan, Joty, and Hoque]{masry-etal-2022-chartqa}
Ahmed Masry, Xuan~Long Do, Jia~Qing Tan, Shafiq Joty, and Enamul Hoque.
\newblock Chartqa: A benchmark for question answering about charts with visual and logical reasoning.
\newblock In \emph{ACL}, pages 2263--2279, 2022{\natexlab{b}}.

\bibitem[Kembhavi et~al.(2016{\natexlab{b}})Kembhavi, Salvato, Kolve, Seo, Hajishirzi, and Farhadi]{kembhavi2016diagram}
Aniruddha Kembhavi, Mike Salvato, Eric Kolve, Minjoon Seo, Hannaneh Hajishirzi, and Ali Farhadi.
\newblock A diagram is worth a dozen images.
\newblock In \emph{ECCV}, pages 235--251. Springer, 2016{\natexlab{b}}.

\bibitem[Cao and Xiao(2022{\natexlab{b}})]{cao-xiao-2022-augmented}
Jie Cao and Jing Xiao.
\newblock An augmented benchmark dataset for geometric question answering through dual parallel text encoding.
\newblock In \emph{COLING}, pages 1511--1520, 2022{\natexlab{b}}.

\bibitem[Kim et~al.(2022{\natexlab{b}})Kim, Hong, Yim, Nam, Park, Yim, Hwang, Yun, Han, and Park]{kim2022donut}
Geewook Kim, Teakgyu Hong, Moonbin Yim, JeongYeon Nam, Jinyoung Park, Jinyeong Yim, Wonseok Hwang, Sangdoo Yun, Dongyoon Han, and Seunghyun Park.
\newblock Ocr-free document understanding transformer.
\newblock In \emph{ECCV}, 2022{\natexlab{b}}.

\bibitem[Li and Lu(2024)]{li2024driving}
Jiajhan Li and Tong Lu.
\newblock Driving with internvl.
\newblock 2024.

\bibitem[Zhang et~al.(2024{\natexlab{b}})Zhang, Zhang, Tian, Fu, Zhang, Wu, Li, Wang, Wen, Zhang, et~al.]{zhang2024mme}
Yi-Fan Zhang, Huanyu Zhang, Haochen Tian, Chaoyou Fu, Shuangqing Zhang, Junfei Wu, Feng Li, Kun Wang, Qingsong Wen, Zhang Zhang, et~al.
\newblock Mme-realworld: Could your multimodal llm challenge high-resolution real-world scenarios that are difficult for humans?
\newblock \emph{arXiv preprint arXiv:2408.13257}, 2024{\natexlab{b}}.

\bibitem[{Anthropic}(2024{\natexlab{b}})]{claude3series2024}
{Anthropic}.
\newblock The claude 3 model family: Opus, sonnet, haiku.
\newblock \url{https://www.anthropic.com}, 2024{\natexlab{b}}.
\newblock URL \url{https://www-cdn.anthropic.com/de8ba9b01c9ab7cbabf5c33b80b7bbc618857627/Model_Card_Claude_3.pdf}.

\bibitem[Kirillov et~al.(2023{\natexlab{b}})Kirillov, Mintun, Ravi, Mao, Rolland, Gustafson, Xiao, Whitehead, Berg, Lo, et~al.]{kirillov2023segment}
Alexander Kirillov, Eric Mintun, Nikhila Ravi, Hanzi Mao, Chloe Rolland, Laura Gustafson, Tete Xiao, Spencer Whitehead, Alexander~C Berg, Wan-Yen Lo, et~al.
\newblock Segment anything.
\newblock In \emph{ICCV}, pages 4015--4026, 2023{\natexlab{b}}.

\bibitem[Kim et~al.(2018)Kim, Rohrbach, Darrell, Canny, and Akata]{kim2018textual}
Jinkyu Kim, Anna Rohrbach, Trevor Darrell, John Canny, and Zeynep Akata.
\newblock Textual explanations for self-driving vehicles.
\newblock \emph{ECCV}, 2018.

\bibitem[Jin et~al.(2023)Jin, Liu, Zheng, Li, Zhao, Zhang, Zheng, Zhou, and Liu]{jin2023adapt}
Bu~Jin, Xinyu Liu, Yupeng Zheng, Pengfei Li, Hao Zhao, Tong Zhang, Yuhang Zheng, Guyue Zhou, and Jingjing Liu.
\newblock Adapt: Action-aware driving caption transformer.
\newblock \emph{arXiv preprint arXiv:2302.00673}, 2023.

\bibitem[Lin et~al.(2023)Lin, Zhao, Zhang, Wu, Zhang, Wang, and Xie]{lin2023pmc}
Weixiong Lin, Ziheng Zhao, Xiaoman Zhang, Chaoyi Wu, Ya~Zhang, Yanfeng Wang, and Weidi Xie.
\newblock Pmc-clip: Contrastive language-image pre-training using biomedical documents.
\newblock \emph{arXiv preprint arXiv:2303.07240}, 2023.

\bibitem[Subramanian et~al.(2020)Subramanian, Wang, Mehta, Bogin, van Zuylen, Parasa, Singh, Gardner, and Hajishirzi]{subramanian-2020-medicat}
Sanjay Subramanian, Lucy~Lu Wang, Sachin Mehta, Ben Bogin, Madeleine van Zuylen, Sravanthi Parasa, Sameer Singh, Matt Gardner, and Hannaneh Hajishirzi.
\newblock Medicat: A dataset of medical images, captions, and textual references.
\newblock In \emph{EMNLP}, 2020.

\bibitem[Wu et~al.(2023{\natexlab{b}})Wu, Zhang, Zhang, Wang, and Xie]{wu2023towards}
Chaoyi Wu, Xiaoman Zhang, Ya~Zhang, Yanfeng Wang, and Weidi Xie.
\newblock Towards generalist foundation model for radiology.
\newblock \emph{arXiv preprint arXiv:2308.02463}, 2023{\natexlab{b}}.

\bibitem[ope()]{openi}
\url{https://openi.nlm.nih.gov/}.

\bibitem[med()]{medpix}
\url{https://medpix.nlm.nih.gov/home}.

\bibitem[Ikezogwo et~al.(2023)Ikezogwo, Seyfioglu, Ghezloo, Geva, Mohammed, Anand, Krishna, and Shapiro]{ikezogwo2023quilt}
Wisdom~Oluchi Ikezogwo, Mehmet~Saygin Seyfioglu, Fatemeh Ghezloo, Dylan Stefan~Chan Geva, Fatwir~Sheikh Mohammed, Pavan~Kumar Anand, Ranjay Krishna, and Linda Shapiro.
\newblock Quilt-1m: One million image-text pairs for histopathology.
\newblock \emph{arXiv preprint arXiv:2306.11207}, 2023.

\bibitem[Johnson et~al.(2018)Johnson, Stone, Celi, and Pollard]{johnson2018mimic}
Alistair E~W Johnson, David~J Stone, Leo~A Celi, and Tom~J Pollard.
\newblock The mimic code repository: enabling reproducibility in critical care research.
\newblock \emph{JAMIA}, 25\penalty0 (1):\penalty0 32--39, 2018.

\bibitem[ima()]{imagebank}
\url{https://imagebank.asrs.org/}.

\bibitem[Liu et~al.(2024{\natexlab{d}})Liu, Li, Li, and Lee]{liu2023improved}
Haotian Liu, Chunyuan Li, Yuheng Li, and Yong~Jae Lee.
\newblock Improved baselines with visual instruction tuning.
\newblock In \emph{CVPR}, pages 26296--26306, 2024{\natexlab{d}}.

\bibitem[Chen et~al.(2024{\natexlab{c}})Chen, Ye, Wang, Li, Deng, Li, Li, Duan, Huang, Su, et~al.]{chen2024gmai}
Pengcheng Chen, Jin Ye, Guoan Wang, Yanjun Li, Zhongying Deng, Wei Li, Tianbin Li, Haodong Duan, Ziyan Huang, Yanzhou Su, et~al.
\newblock Gmai-mmbench: A comprehensive multimodal evaluation benchmark towards general medical ai.
\newblock \emph{arXiv preprint arXiv:2408.03361}, 2024{\natexlab{c}}.

\bibitem[Luo et~al.(2024{\natexlab{b}})Luo, Pang, Zhang, Wang, Wang, Dang, Lao, Wang, Chen, Tan, and Li]{luo2024sky}
Junwei Luo, Zhen Pang, Yongjun Zhang, Tingzhu Wang, Linlin Wang, Bo~Dang, Jiangwei Lao, Jian Wang, Jingdong Chen, Yihua Tan, and Yansheng Li.
\newblock Skysensegpt: A fine-grained instruction tuning dataset and model for remote sensing vision-language understanding.
\newblock \emph{arXiv preprint arXiv:2406.10100}, 2024{\natexlab{b}}.

\bibitem[Zhan et~al.(2023)Zhan, Xiong, and Yuan]{zhan2023rsvg}
Yang Zhan, Zhitong Xiong, and Yuan Yuan.
\newblock Rsvg: Exploring data and models for visual grounding on remote sensing data.
\newblock \emph{TGRS}, 61:\penalty0 1--13, 2023.

\bibitem[Zhang et~al.(2023{\natexlab{c}})Zhang, Jiao, Li, Liu, Chen, Liu, Li, and Guo]{zhang2023spatial}
Zixiao Zhang, Licheng Jiao, Lingling Li, Xu~Liu, Puhua Chen, Fang Liu, Yuxuan Li, and Zhicheng Guo.
\newblock A spatial hierarchical reasoning network for remote sensing visual question answering.
\newblock \emph{TGRS}, 61:\penalty0 1--15, 2023{\natexlab{c}}.

\bibitem[Bazi et~al.(2022)Bazi, Al~Rahhal, Mekhalfi, Al~Zuair, and Melgani]{bazi2022bi}
Yakoub Bazi, Mohamad~Mahmoud Al~Rahhal, Mohamed~Lamine Mekhalfi, Mansour~Abdulaziz Al~Zuair, and Farid Melgani.
\newblock Bi-modal transformer-based approach for visual question answering in remote sensing imagery.
\newblock \emph{TGRS}, 60:\penalty0 1--11, 2022.

\bibitem[Yuan et~al.(2022)Yuan, Mou, Wang, and Zhu]{yuan2022easy}
Zhenghang Yuan, Lichao Mou, Qi~Wang, and Xiao~Xiang Zhu.
\newblock From easy to hard: Learning language-guided curriculum for visual question answering on remote sensing data.
\newblock \emph{TGRS}, 60:\penalty0 1--11, 2022.

\bibitem[Zhan et~al.(2024)Zhan, Xiong, and Yuan]{zhan2024skyeyegpt}
Yang Zhan, Zhitong Xiong, and Yuan Yuan.
\newblock Skyeyegpt: Unifying remote sensing vision-language tasks via instruction tuning with large language model.
\newblock \emph{arXiv preprint arXiv:2401.09712}, 2024.

\bibitem[Rajbhandari et~al.(2020)Rajbhandari, Rasley, Ruwase, and He]{rajbhandari2020zero}
Samyam Rajbhandari, Jeff Rasley, Olatunji Ruwase, and Yuxiong He.
\newblock Zero: Memory optimizations toward training trillion parameter models.
\newblock In \emph{SC20: International Conference for High Performance Computing, Networking, Storage and Analysis}, pages 1--16. IEEE, 2020.

\end{thebibliography}
\end{document}